\documentclass{article}

\usepackage{iclr2027_conference}

\usepackage{times}
\usepackage{graphicx}
\usepackage{amsmath}
\usepackage{amssymb}
\usepackage{booktabs}
\usepackage{multirow}
\usepackage{algorithm}
\usepackage{algorithmic}
\usepackage{hyperref}
\usepackage{url}
\usepackage{float}
\usepackage{caption}

\usepackage[table]{xcolor}

\definecolor{bestblue}{RGB}{196,210,243}
\definecolor{resourcegreen}{RGB}{70,145,55}

\definecolor{projectlink}{RGB}{75,0,130} 
\newcommand{\projecthref}[2]{%
  \begingroup
  \hypersetup{hidelinks}
  \href{#1}{\textcolor{projectlink}{\texttt{#2}}}%
  \endgroup
}

\usepackage{fancyhdr}
\fancypagestyle{preprintfirst}{%
  \fancyhf{}
  \fancyhead[L]{%
    \IfFileExists{figures/BAAI_brand.png}{%
      \includegraphics[
        height=12pt,
        keepaspectratio
      ]{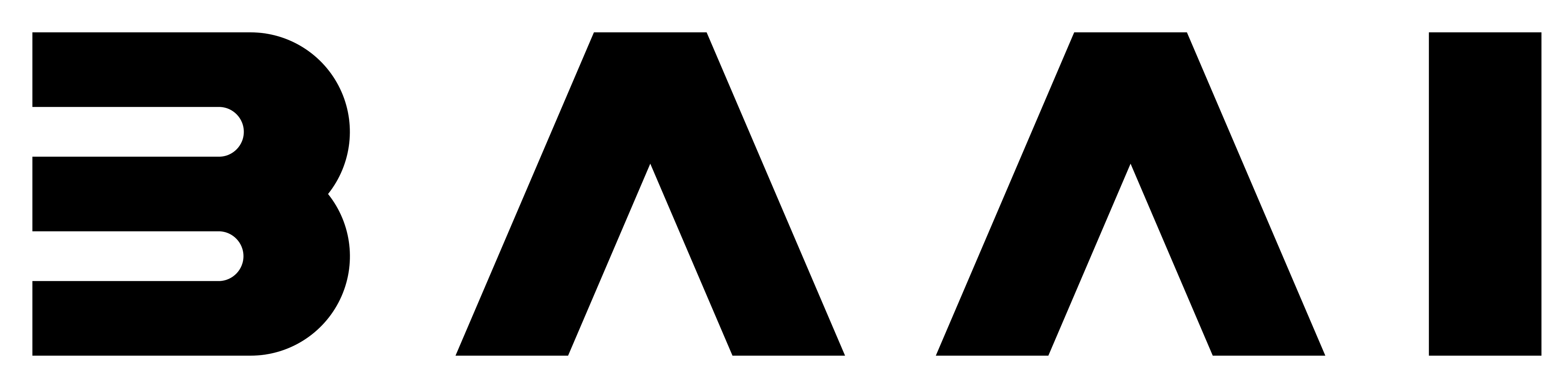}%
    }{\small\textbf{BAAI}}%
  }
  \fancyfoot[C]{\thepage}
  \renewcommand{\headrulewidth}{0.4pt}
}

\fancypagestyle{preprintrest}{%
  \fancyhf{}
  \fancyhead[L]{%
    \IfFileExists{figures/BAAI_brand.png}{%
      \includegraphics[
        height=12pt,
        keepaspectratio
      ]{figures/BAAI_brand.png}%
    }{\small\textbf{BAAI}}%
  }
  \fancyfoot[C]{\thepage}
  \renewcommand{\headrulewidth}{0.4pt}
}

\title{
MOST: Memory of Object States and\\
Task Progress for Long-Horizon Mobile\\
Manipulation
}

\newcommand{\paperauthor}[2]{%
  \textbf{#1}\textsuperscript{\textbf{#2}}%
}

\author{
\begin{tabular}{@{}l@{\quad}l@{\quad}l@{\quad}l@{}}
\paperauthor{Diqi He}{1,2} &
\paperauthor{Jinbang Guo}{2} &
\paperauthor{Yijun Shen}{2,3} &
\paperauthor{Zhiwei Yu}{2,*} \\
\paperauthor{Huan Yang}{2} &
\paperauthor{Dingwen Zhang}{1,*} &
\paperauthor{Junwei Han}{4} &
\paperauthor{Zhongyuan Wang}{2}
\end{tabular}
\\[4pt]
{\normalfont\small
\textsuperscript{1}Northwestern Polytechnical University
\qquad
\textsuperscript{2}Beijing Academy of Artificial Intelligence
}\\
{\normalfont\small
\textsuperscript{3}East China Normal University
\qquad
\textsuperscript{4}Chongqing University of Posts and Telecommunications
}\\
{\normalfont\small
\textsuperscript{*}Corresponding authors.
}\\[10pt]
{\normalfont\small \textbf{Project Page:} \projecthref{https://most-vla.github.io/}{https://most-vla.github.io/}}\\
{\normalfont\small \textbf{MOST Code:} \projecthref{https://github.com/diqihe666/MOST-vla}{https://github.com/diqihe666/MOST-vla}}\\
{\normalfont\small \textbf{Data:} \projecthref{https://huggingface.co/datasets/diqihe666/robocasa_long}{https://huggingface.co/datasets/diqihe666/robocasa\_long}}
}

\iclrfinalcopy

\begin{document}

\maketitle
\pagestyle{preprintrest}
\thispagestyle{preprintfirst}

\begin{figure}[H]
  \centering
  \captionsetup{skip=5pt}
  \includegraphics[width=\linewidth]{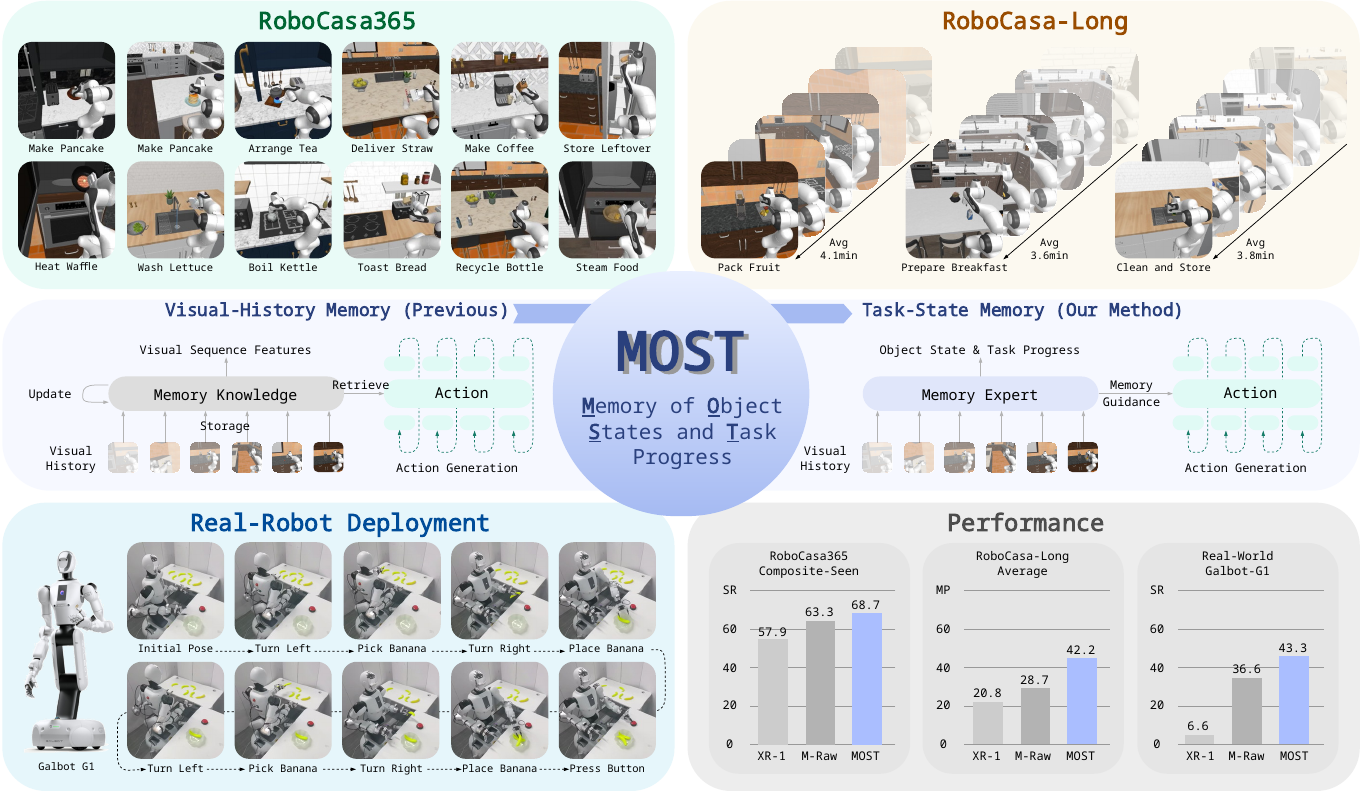}
  \caption{\textbf{MOST} structures policy-internal memory around object states and task progress for long-horizon mobile manipulation. We evaluate it on RoboCasa365, our newly introduced \textbf{RoboCasa-Long} benchmark of multi-minute, multi-stage tasks, and real-world deployment. M-Raw in the plots denotes MOST-Raw, a matched visual-history memory variant without task-state supervision or future prediction. Across these settings, MOST outperforms both the memoryless Xiaomi-Robotics-1 (XR-1) baseline and MOST-Raw.}
  \label{fig:teaser}
\end{figure}


\begin{abstract}
Long-horizon embodied manipulation requires policies to act on information that is no longer visible. Modified objects may leave the field of view, while completed subtasks continue to constrain future actions. Existing approaches extend temporal context through recurrent state updates, test-time training, or storage and retrieval of past observations. As horizons grow, these mechanisms can still leave the policy to infer task state from long, redundant, and viewpoint-dependent visual experience. We argue that long-horizon agents benefit not merely from remembering more, but from remembering task-relevant state changes and progress. We propose \textbf{Memory of Object States and Task Progress (MOST)}, a policy-internal memory framework that compresses multi-view observations into temporal tokens. Historical and predictive future tokens are explicitly supervised with object states and subtask progress, structuring memory around what has changed in the world and what has already been accomplished. The resulting memory directly conditions action generation within a single end-to-end policy, without an external planner or retrieval stage. We further introduce \textbf{RoboCasa-Long}, a benchmark of multi-minute, multi-stage mobile manipulation tasks. Across RoboCasa365, RoboCasa-Long, and real-world robot experiments, MOST consistently outperforms Xiaomi-Robotics-1 (XR-1), the previous state-of-the-art model. It raises average success from 65.9\% to 72.1\% on RoboCasa365, average maximum task progress from 20.83\% to 42.22\% on RoboCasa-Long, and real-world success from 6.66\% to 43.33\%. MOST also outperforms a matched visual-history memory variant, showing that effective long-horizon memory depends not only on how much experience is retained, but on whether it preserves decision-relevant progress and world-state changes.

\end{abstract}


\section{Introduction}
Learning-based robot policies are increasingly capable across tasks, objects, instructions, and embodiments~\citep{openx2023,kim2024openvla}. Scaling robot data, multimodal representations, and action models has broadened task coverage and improved dexterous control~\citep{black2025pi05,chi2023diffusion}. Yet extending these policies to long-horizon embodied tasks introduces a qualitatively different challenge. As a robot moves across workspaces and executes multiple task stages, information that remains critical for future decisions can disappear from the current observation~\citep{dai2026robomme,yang2026eventvla}. An object may already have been moved, an appliance may have been turned on, or a subtask may have been completed several minutes earlier. The difficulty is therefore not only that the current observation is partial. The policy must preserve the task-relevant consequences of earlier interactions, particularly persistent object states and task progress. Losing this information can lead to repeated actions, skipped stages, incorrect object interactions, or premature termination~\citep{bagaria2026rbvla}. These dependencies are especially relevant in mobile, multi-workspace settings~\citep{yenamandra2023homerobot,fu2024mobilealoha}.

Existing approaches extend temporal context through recurrent or compressed state updates, test-time training, and storage or retrieval of past observations~\citep{cherepanov2026muvla,sun2025ttt,shi2025memoryvla,shah2026halo}. Despite their different mechanisms, these approaches largely focus on how much history to retain and how it should be updated or accessed. Progress-critical events can nevertheless remain embedded in long visual streams, while visually similar stages may require different next actions. We instead focus on what information from interaction history should survive because it remains relevant for future decisions. This distinction defines our controlled comparison among a memoryless policy, a matched visual-history memory, and memory explicitly structured by task state.

To realize this structured alternative, we organize a single memory sequence around two complementary kinds of task-relevant information: task progress and object states. Task progress records the current phase and what has already been completed~\citep{bagaria2026rbvla}, while object states record action-induced changes to relevant objects that may no longer be visible~\citep{chung2026slotssm}. Their temporal evolution links earlier interactions to future decisions while suppressing visual detail that is no longer relevant. For example, packing three bananas requires tracking completed transfers and whether the completion button should be pressed after earlier bananas leave the current view.

Based on this representation, we propose \textbf{Memory of Object States and Task Progress (MOST)}, an instruction-conditioned, end-to-end policy with compact task-state memory. MOST compresses multi-view observations into temporal tokens and structures both historical and predictive tokens through subtask-progress and relevant object-state supervision. This memory directly conditions action generation, while the auxiliary supervision is removed at inference. MOST therefore operates as a single policy without an external planner or retrieval stage.

Evaluating this hypothesis requires tasks in which relevant state persists over substantially longer horizons than in standard manipulation benchmarks. RoboCasa365~\citep{nasiriany2026robocasa365} offers broad task coverage, but its standard episodes place limited pressure on memory over multi-minute execution. Existing memory-focused benchmarks introduce explicit long-term dependencies, often in constrained or purpose-built settings~\citep{dai2026robomme,chen2026rmbench}. This leaves room for natural household activities such as preparing food, cleaning, and tidying that span multiple work areas. We therefore introduce \textbf{RoboCasa-Long}, a benchmark of multi-minute, multi-stage mobile manipulation tasks that require policies to preserve task progress and world-state changes across extended navigation and interaction. As summarized in Fig.~\ref{fig:teaser}, we evaluate MOST on RoboCasa365, RoboCasa-Long, and real-world deployment. We compare against the previous state-of-the-art Xiaomi-Robotics-1 (XR-1) baseline~\citep{fan2025xr1} and a matched visual-history memory variant. Progress-aware metrics adapted from PRM-as-a-Judge 1.5~\citep{liu2026prmjudge} and controlled representation studies separate access to history from how it is represented. MOST raises average success from 65.9\% to 72.1\% on RoboCasa365, average maximum task progress from 20.83\% to 42.22\% on RoboCasa-Long, and real-world success from 6.66\% to 43.33\%. These results support structuring temporal context around task progress and object states.

Our contributions are threefold:
\begin{itemize}
    \item \textbf{Memory formulation.} We formulate long-horizon policy memory as a representation problem centered on what task-relevant information should survive, rather than only how much history is retained or how it is accessed.
    \item \textbf{MOST.} We introduce a policy-internal memory framework that learns compact temporal representations under subtask-progress and object-state supervision, directly conditioning action generation without external planning or retrieval.
    \item \textbf{Long-horizon evaluation.} We introduce \textbf{RoboCasa-Long}, a benchmark of multi-minute, multi-stage mobile manipulation, and demonstrate consistent gains across RoboCasa365, RoboCasa-Long, and real-world deployment.
\end{itemize}

\section{Related Work}
\subsection{Memory for Long-Horizon Robot Control}

Long-horizon robot control requires policies to preserve information after relevant evidence leaves the current view. Existing methods extend temporal context through recurrence and attention~\citep{parisotto2020gtrxl,cherepanov2026muvla,bagaria2026rbvla}, test-time training~\citep{sun2025ttt,jiang2026robottt}, compressed multi-frame memory~\citep{jang2025contextvla,torne2026mem,shah2026prism,wang2026nativemem}, or event selection and retrieval~\citep{yang2026eventvla,shah2026halo}. Structured maps provide another route by organizing past observations spatially~\citep{parisotto2018neural,chaplot2020neural}. Hierarchical systems instead externalize long-horizon reasoning through language-model plans or executable programs~\citep{ahn2022saycan,huang2022inner,liang2023code}. These approaches establish several effective ways to preserve and access longer context. MOST complements them by shifting the focus from how history is retained or accessed to what information should persist in policy memory.

\subsection{Task-State Representations for Long-Horizon Manipulation}

For extended manipulation, useful memory should retain the consequences of earlier interactions even when their visual evidence is no longer available. Recurrent models summarize latent execution context~\citep{cherepanov2026muvla,bagaria2026rbvla}, while perceptual-cognitive and imagination-based memories integrate prior observations with predicted future context~\citep{shi2025memoryvla,shi2026memoryvlapp}. Object-centric memory further organizes temporal state around persistent entities~\citep{chung2026slotssm}. High-level planners can also express task state through intermediate plans, feedback, or executable steps~\citep{ahn2022saycan,huang2022inner,singh2023progprompt}, although this state is maintained outside the action policy. Building on these related directions, MOST jointly structures policy-internal memory around subtask progress and the states of relevant objects. The resulting representation preserves both what has been accomplished and how earlier actions changed the world.

\subsection{Long-Horizon Manipulation Benchmarks}

Evaluating long-horizon memory requires tasks in which earlier events remain relevant throughout extended execution. CALVIN and HULC evaluate language-conditioned skill sequences~\citep{mees2021calvin,mees2022hulc}, while LIBERO studies transfer across manipulation task suites~\citep{liu2023libero}. ALFRED, RoboCasa, and RoboCasa365 broaden evaluation to household activities with multiple objects, locations, and state changes~\citep{shridhar2020alfred,nasiriany2024robocasa,nasiriany2026robocasa365}. Mobile manipulation systems additionally demonstrate operation across work areas~\citep{yenamandra2023homerobot,fu2024mobilealoha,wu2023tidybot,liu2024okrobot}. These settings contain important sequential structure, but are not designed specifically to isolate persistent task state throughout multi-minute interaction across several workspaces. RoboMME and RMBench directly probe memory-dependent manipulation through controlled task designs~\citep{dai2026robomme,chen2026rmbench}. Complementing both lines, RoboCasa-Long evaluates natural multi-stage household activities whose progress and world-state changes must remain available as the robot moves among work areas. It therefore provides a focused testbed for the memory representation studied by MOST.

\section{Method}
\begin{figure}[t]
    \centering
    \captionsetup{skip=5pt}
    \includegraphics[width=\linewidth]{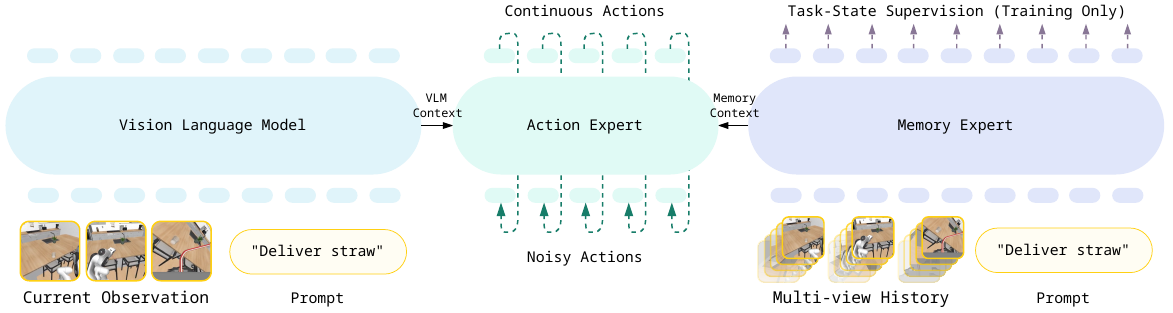}
    \caption{\textbf{MOST overview.} MOST integrates a Vision Language Model, a Memory Expert, and an Action Expert within one end-to-end policy. The Vision Language Model encodes the current observation and instruction. The Memory Expert represents sampled history as object states and task progress, and the Action Expert uses both contexts to generate continuous actions. Task-state supervision is training-only.}
    \label{fig:most_overview}
\end{figure}

\subsection{MOST Overview}

MOST is an instruction-conditioned, end-to-end policy with three components (Fig.~\ref{fig:most_overview}). The \textbf{Vision Language Model} encodes the current multi-view observation and instruction. The \textbf{Memory Expert} converts sampled history into an instruction-conditioned representation of object states and task progress. The \textbf{Action Expert} uses both contexts to predict continuous actions. The two experts are depth aligned, letting task-state information influence action generation without an external planner or retrieval module.

\subsection{Compact Task-State Memory}

\begin{figure}[t]
    \centering
    \captionsetup{skip=5pt}
    \includegraphics[width=\linewidth]{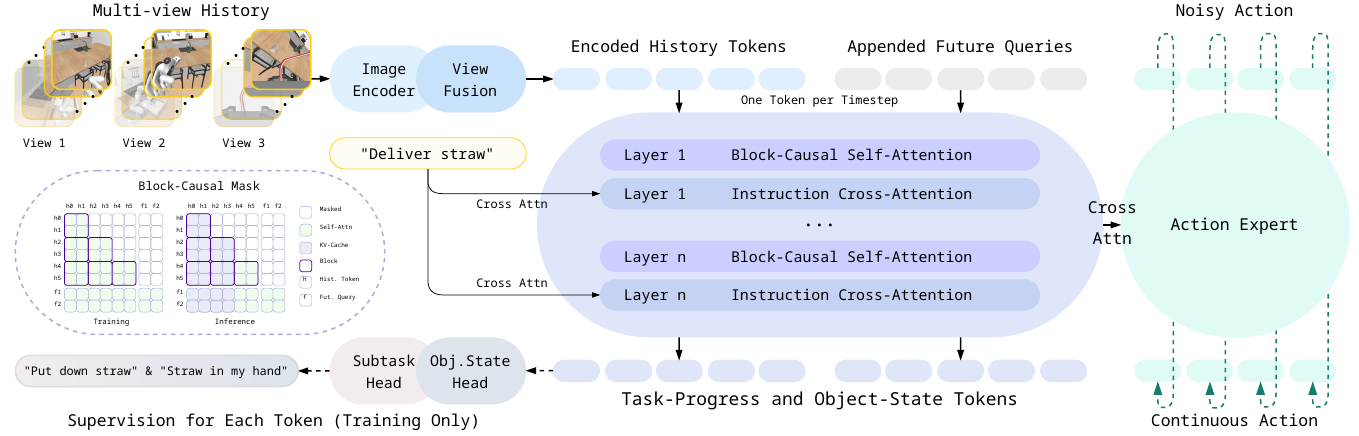}
    \caption{\textbf{Memory Expert architecture.} Historical multi-view observations are encoded into history tokens, and future queries are appended to form the memory sequence. The sequence is processed with block-causal self-attention and conditioned on the instruction through cross-attention. During training, each token predicts its subtask and the state of every relevant object. The output of each Memory Expert layer conditions the corresponding Action Expert layer.}
    \label{fig:most_pipeline}
\end{figure}

Fig.~\ref{fig:most_pipeline} details the Memory Expert. At sampled timestep $i$, the multi-view observation is $O_i=\{I_i^{(v)}\}_{v=1}^{V}$, where $I_i^{(v)}$ is the image from view $v$ among $V$ views. A DINOv2 image encoder~\citep{oquab2023dinov2} extracts each view's global $[\mathrm{CLS}]$ feature $c_i^{(v)}$. A learned view query $q_{\mathrm{view}}$ cross-attends to these features to produce one history token $h_i$. One token per sampled timestep makes memory growth independent of views and image patches, keeping long histories computationally manageable.

To make the compressed memory decision-relevant, we structure it to preserve observed task progress while also encoding anticipated future task states. We append $K$ copies of a learned future query $q_{\mathrm{fut}}$ to the $N$ history tokens. Let $\ell$ denote the encoded instruction and $E_{\mathrm{mem}}$ the Memory Expert. The history tokens and complete memory sequence are
\begin{equation}
\begin{gathered}
h_i = \operatorname{ViewFuse}\!\left(q_{\mathrm{view}},[c_i^{(1)},\ldots,c_i^{(V)}]\right),\\
M=[m_1,\ldots,m_{N+K}]
=E_{\mathrm{mem}}\!\left(
[h_1,\ldots,h_N,
\underbrace{q_{\mathrm{fut}},\ldots,q_{\mathrm{fut}}}_{K\ \mathrm{future\ queries}}],\ell
\right).
\end{gathered}
\label{eq:memory_sequence}
\end{equation}
Here, $\operatorname{ViewFuse}$ is learned-query cross-attention over the views. The output sequence $M$ contains a memory token $m_i$ at position $i$.

The Memory Expert groups the history into consecutive blocks of $B$ tokens. Block-causal self-attention is bidirectional within a block and causal across blocks, so history tokens cannot access later blocks or future queries. Future queries attend to the full history and one another. Each layer then cross-attends to the encoded instruction $\ell$, conditioning memory on the commanded task. During inference, completed blocks are cached while only the current block and future queries are updated. This maintains ordered context without reprocessing the full history.

\subsection{Memory-Conditioned Action Generation}

Each Action Expert layer attends to its paired Memory Expert output, making task-state memory directly available during control (Fig.~\ref{fig:most_pipeline}). At layer $r$, let $M^{(r)}$ and $X^{(r)}$ denote the memory and action features. The action features query the memory through gated cross-attention:
\begin{equation}
\begin{gathered}
Q^{(r)}=\operatorname{Query}\!\left(X^{(r)}\right),\qquad
K^{(r)}=\operatorname{Key}\!\left(M^{(r)}\right),\qquad
V^{(r)}=\operatorname{Value}\!\left(M^{(r)}\right),\\
\widetilde{X}^{(r)}=X^{(r)}+g^{(r)}\!\cdot\!
\operatorname{Attention}\!\left(Q^{(r)},K^{(r)},V^{(r)}\right).
\end{gathered}
\label{eq:memory_action}
\end{equation}
Here, $\widetilde{X}^{(r)}$ is the memory-conditioned action representation and $g^{(r)}$ is a learned scalar gate initialized to zero. Applying this interaction at every paired layer exposes action generation to observed progress and anticipated task states throughout the network.

\subsection{Task-Progress and Object-State Supervision}

We supervise memory with subtask and object-state targets. At position $i$, label $s_i$ identifies the execution stage, such as picking up or placing an object. For each relevant object $o$, label $z_{i,o}$ describes its state, such as being on a counter, in hand, or at the destination. We combine its category and episode-local instance embeddings into $e_o$, allowing the state head to distinguish instances.

Each memory output $m_i$ predicts one subtask and one state for every relevant object. The subtask head $H_{\mathrm{sub}}$ receives $m_i$, while the shared object-state head $H_{\mathrm{obj}}$ also receives $e_o$. Thus, a task with three relevant objects gives each token one subtask target and three object-state targets. Let $\mathcal{O}$ be the set of relevant objects. The cross-entropy objectives are
\begin{equation}
\begin{gathered}
\mathcal{L}_{\mathrm{sub}}
=\operatorname*{mean}_{i}\operatorname{CE}\!\left(H_{\mathrm{sub}}(m_i),s_i\right),\qquad
\mathcal{L}_{\mathrm{obj}}
=\operatorname*{mean}_{i,\,o\in\mathcal{O}}\operatorname{CE}\!\left(H_{\mathrm{obj}}(m_i,e_o),z_{i,o}\right),\\
\mathcal{L}
=\mathcal{L}_{\mathrm{policy}}
+\lambda_{\mathrm{sub}}\mathcal{L}_{\mathrm{sub}}
+\lambda_{\mathrm{obj}}\mathcal{L}_{\mathrm{obj}}.
\end{gathered}
\label{eq:training_objective}
\end{equation}
The means cover the available annotations in each training sample. Here, $\mathcal{L}_{\mathrm{policy}}$ is the action-prediction objective, and $\lambda_{\mathrm{sub}}/\lambda_{\mathrm{obj}}$ denotes the auxiliary loss weight. The subtask and object-state vocabularies are closed-set and constructed from each training split. The auxiliary heads and object embeddings are removed at inference, while the Memory Expert, future queries, and memory-to-action pathway remain active. Further details appear in Appendix~\ref{app:implementation}.

\section{Experiments}
We test whether representing policy memory through object states and task progress improves control beyond unstructured visual history. We report success and execution progress on RoboCasa365, multi-minute mobile manipulation in simulation, and real-world count-conditioned manipulation.

\begin{table}[t]
\captionsetup{skip=4pt}
\centering
\caption{
Success rate (\%) on the 50-task RoboCasa365 target suite under Target-10\% and Target-100\%.
GR00T N1.5 split rates are published results, while XR-1, MOST-Raw, and MOST use our matched protocol.
Averages are recomputed from the displayed split rates with 18/16/16 task weighting and truncated to one decimal place.
The best result in each setting is shown in \textbf{bold}.
}
\label{tab:robocasa365_main}

\setlength{\tabcolsep}{7.0pt}
\renewcommand{\arraystretch}{0.6}

\begin{tabular}{llcccc}
\toprule
Data & Method
& Atomic$\uparrow$
& Composite-S$\uparrow$
& Composite-U$\uparrow$
& Average$\uparrow$ \\
\midrule

\multirow{4}{*}{Target 10\%}
& GR00T N1.5
& 57.0 & 28.7 & 32.3 & 40.0 \\

& XR-1
& \cellcolor{bestblue!33}60.1 
& \cellcolor{bestblue!33}44.3 
& \cellcolor{bestblue!33}43.1 
& \cellcolor{bestblue!33}49.6 \\

\cmidrule(lr){2-6}
& MOST-Raw
& \cellcolor{bestblue!66}61.1 
& \cellcolor{bestblue!66}48.3 
& \cellcolor{bestblue!66}45.2 
& \cellcolor{bestblue!66}51.9 \\

& \textbf{MOST (Ours)}
& \cellcolor{bestblue}\textbf{63.3}
& \cellcolor{bestblue}\textbf{51.0}
& \cellcolor{bestblue}\textbf{48.7}
& \cellcolor{bestblue}\textbf{54.6} \\

\midrule

\multirow{4}{*}{Target 100\%}
& GR00T N1.5
& 70.0 & 41.2 & 44.0 & 52.4 \\

& XR-1
& \cellcolor{bestblue!66}78.5 
& \cellcolor{bestblue!33}57.9 
& \cellcolor{bestblue!33}60.0 
& \cellcolor{bestblue!33}65.9 \\

\cmidrule(lr){2-6}
& MOST-Raw
& \cellcolor{bestblue!33}77.5 
& \cellcolor{bestblue!66}63.3 
& \cellcolor{bestblue!66}62.2 
& \cellcolor{bestblue!66}68.0 \\

& \textbf{MOST (Ours)}
& \cellcolor{bestblue}\textbf{81.8}
& \cellcolor{bestblue}\textbf{68.7}
& \cellcolor{bestblue}\textbf{64.7}
& \cellcolor{bestblue}\textbf{72.1} \\

\bottomrule
\end{tabular}
\end{table}

\begin{table}[t]
\captionsetup{skip=4pt}
\centering
\caption{
Progress-aware evaluation (\%) on RoboCasa365 composite tasks.
XR-1, MOST-Raw, and MOST use matched data, training budgets, and evaluation protocols.
Best results within each setting are highlighted in blue and \textbf{bold}. Second-best results use a lighter blue.
}
\label{tab:robocasa365_progress}

\footnotesize
\setlength{\tabcolsep}{4.0pt}
\renewcommand{\arraystretch}{0.6}

\resizebox{\linewidth}{!}{%
\begin{tabular}{llcccccccc}
\toprule
Data & Method
& SR$\uparrow$
& MC@25$\uparrow$
& MC@50$\uparrow$
& MC@75$\uparrow$
& MP$\uparrow$
& PPL$\uparrow$
& OPS$\uparrow$
& UPS$\uparrow$ \\
\midrule

\multirow{3}{*}{Com-S-10\%}
& XR-1
& 44.37
& \cellcolor{bestblue!35}88.12
& 77.29
& 61.45
& 72.34
& 57.64
& 66.42
& 79.22 \\

& MOST-Raw
& \cellcolor{bestblue!35}48.33
& 86.45
& \cellcolor{bestblue!35}77.70
& \cellcolor{bestblue!35}64.16
& \cellcolor{bestblue!35}73.78
& \cellcolor{bestblue!35}58.54
& \cellcolor{bestblue!35}67.12
& \cellcolor{bestblue!35}79.46 \\

& \textbf{MOST (Ours)}
& \cellcolor{bestblue}\textbf{51.04}
& \cellcolor{bestblue}\textbf{89.16}
& \cellcolor{bestblue}\textbf{82.50}
& \cellcolor{bestblue}\textbf{68.75}
& \cellcolor{bestblue}\textbf{77.29}
& \cellcolor{bestblue}\textbf{62.00}
& \cellcolor{bestblue}\textbf{71.86}
& \cellcolor{bestblue}\textbf{81.76} \\

\midrule

\multirow{3}{*}{Com-U-10\%}
& XR-1
& 43.12
& \cellcolor{bestblue!35}94.37
& 80.83
& 58.75
& 73.30
& 56.02
& 66.92
& \cellcolor{bestblue!35}81.08 \\

& MOST-Raw
& \cellcolor{bestblue!35}45.20
& 94.16
& \cellcolor{bestblue!35}81.66
& \cellcolor{bestblue}\textbf{61.04}
& \cellcolor{bestblue!35}74.41
& \cellcolor{bestblue!35}56.99
& \cellcolor{bestblue!35}68.69
& 81.05 \\

& \textbf{MOST (Ours)}
& \cellcolor{bestblue}\textbf{48.75}
& \cellcolor{bestblue}\textbf{95.41}
& \cellcolor{bestblue}\textbf{83.33}
& \cellcolor{bestblue!35}60.83
& \cellcolor{bestblue}\textbf{75.79}
& \cellcolor{bestblue}\textbf{57.55}
& \cellcolor{bestblue}\textbf{69.11}
& \cellcolor{bestblue}\textbf{83.49} \\

\midrule

\multirow{3}{*}{Com-S-100\%}
& XR-1
& 57.91
& \cellcolor{bestblue!35}90.83
& 84.37
& 70.41
& 79.48
& 63.55
& 74.83
& 84.56 \\

& MOST-Raw
& \cellcolor{bestblue!35}63.33
& 90.62
& \cellcolor{bestblue!35}85.00
& \cellcolor{bestblue!35}73.95
& \cellcolor{bestblue!35}81.68
& \cellcolor{bestblue!35}65.06
& \cellcolor{bestblue!35}76.67
& \cellcolor{bestblue!35}86.82 \\

& \textbf{MOST (Ours)}
& \cellcolor{bestblue}\textbf{68.75}
& \cellcolor{bestblue}\textbf{94.79}
& \cellcolor{bestblue}\textbf{89.16}
& \cellcolor{bestblue}\textbf{80.20}
& \cellcolor{bestblue}\textbf{85.88}
& \cellcolor{bestblue}\textbf{68.50}
& \cellcolor{bestblue}\textbf{82.09}
& \cellcolor{bestblue}\textbf{89.37} \\

\midrule

\multirow{3}{*}{Com-U-100\%}
& XR-1
& 60.00
& 95.41
& 87.50
& 73.75
& 82.03
& 64.54
& 76.70
& 88.12 \\

& MOST-Raw
& \cellcolor{bestblue!35}62.29
& \cellcolor{bestblue}\textbf{97.50}
& \cellcolor{bestblue!35}88.95
& \cellcolor{bestblue!35}74.79
& \cellcolor{bestblue!35}83.39
& \cellcolor{bestblue!35}67.16
& \cellcolor{bestblue!35}78.10
& \cellcolor{bestblue}\textbf{89.23} \\

& \textbf{MOST (Ours)}
& \cellcolor{bestblue}\textbf{64.79}
& \cellcolor{bestblue!35}96.45
& \cellcolor{bestblue}\textbf{90.20}
& \cellcolor{bestblue}\textbf{76.25}
& \cellcolor{bestblue}\textbf{84.66}
& \cellcolor{bestblue}\textbf{68.38}
& \cellcolor{bestblue}\textbf{79.71}
& \cellcolor{bestblue!35}89.13 \\

\bottomrule
\end{tabular}%
}

\end{table}

\subsection{Experimental Setup}

\paragraph{Benchmarks and protocol.}
The RoboCasa365 target suite~\citep{nasiriany2026robocasa365} contains 18 Atomic, 16 Composite-Seen, and 16 Composite-Unseen tasks. We first pretrain on Human300 and then finetune on the target suite. Target-10\% and Target-100\% use 50 and 500 target demonstrations per task, respectively, with 60k finetuning steps. Our \textbf{RoboCasa-Long} benchmark contains three multi-minute, multi-stage household tasks across work areas. Each provides 200 demonstrations averaging 3.8\,min, and joint training runs for 20k steps. We evaluate 30 rollouts per task with matched data and schedules. Further details are provided in Appendix~\ref{app:experimental_details}.

\paragraph{Comparisons.}
We report published GR00T N1.5 results~\citep{nasiriany2026robocasa365,groot2025}. XR-1~\citep{fan2025xr1}, MOST-Raw, and MOST use matched data, training budgets, and evaluation. XR-1 is memoryless. \textbf{MOST-Raw} retains visual-history tokens and the memory-to-action interface but removes task-state supervision and future queries, providing an unstructured visual-history baseline.

\paragraph{Metrics.}
Alongside success rate (SR), we adapt Milestone Coverage (MC@25/50/75), Max Progress (MP), and Path-weighted Progress Length (PPL) from PRM-as-a-Judge 1.5~\citep{liu2026prmjudge}. MC measures threshold attainment, MP furthest progress, and PPL path efficiency. Ordered Progress Score (OPS) and Unordered Progress Score (UPS) measure ordered and unordered subtask completion. Simulation uses simulator state, while real trials use the same stages scored from recordings. Appendix~\ref{app:metrics} gives full definitions.

\subsection{RoboCasa365}
\label{sec:robocasa365_results}

RoboCasa365 spans diverse household tasks with multiple objects, state changes, and stages, providing a controlled test of what policy memory should represent. We compare three matched policies. XR-1 is memoryless. MOST-Raw encodes visual history without task-state supervision or future queries, whereas MOST supervises historical and future tokens with subtasks and relevant object states. Tab.~\ref{tab:robocasa365_main} reports SR across all splits and both data regimes. MOST-Raw improves average SR over XR-1 in both regimes. MOST reaches 54.6\% and 72.1\% and leads every split. MOST-Raw's Target-100\% Atomic regression shows that extra context is not uniformly beneficial, supporting memory structured around task state.

To evaluate behavior before terminal success, Tab.~\ref{tab:robocasa365_progress} measures milestone attainment, progress efficiency, and ordered and unordered completion. MOST surpasses MOST-Raw in SR, MP, PPL, and OPS across all four composite settings. Higher PPL and OPS indicate that task-state supervision supports efficient, correctly ordered execution, not only greater unordered completion.

\begin{figure}[t]
    \centering
    \captionsetup{skip=5pt}
    \includegraphics[width=\linewidth]{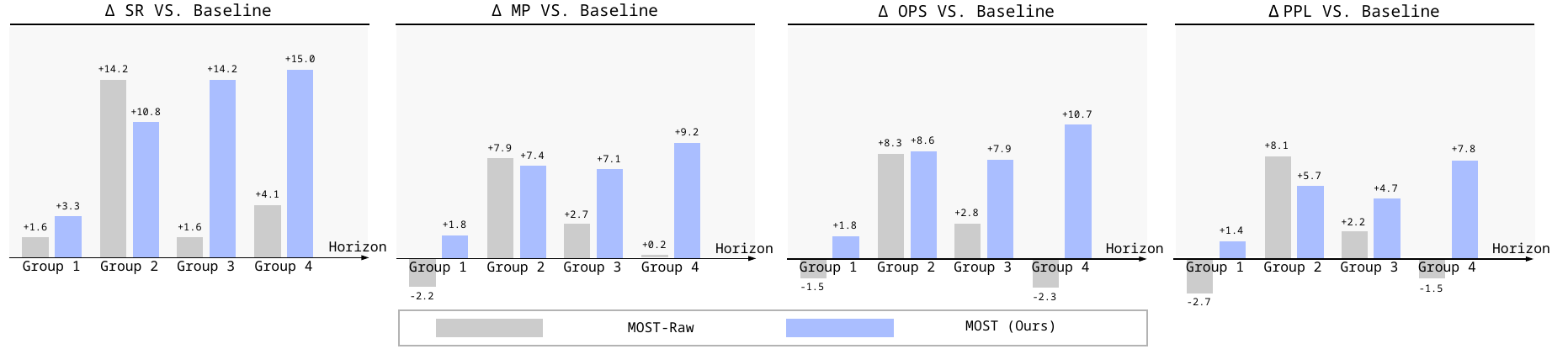}
    \caption{Horizon analysis on RoboCasa365 Composite-Seen under Target-100\%. We sort 16 tasks by demonstration length into four groups from shortest (G1) to longest (G4) and report mean changes from XR-1. MOST separates most clearly from unstructured visual history on G3 and G4.}
    \label{fig:exp_horizon}
\end{figure}

\begin{table}[t]
\captionsetup{skip=4pt}
\centering
\caption{
Evaluation on our RoboCasa-Long benchmark.
We compare XR-1, MOST-Raw, and MOST on three multi-minute, multi-stage manipulation tasks.
Best results, including ties, are highlighted in blue and \textbf{bold}; a distinct second-best result uses lighter blue.
}
\label{tab:robocasa_long}

\footnotesize
\setlength{\tabcolsep}{4.0pt}
\renewcommand{\arraystretch}{0.6}

\resizebox{\linewidth}{!}{%
\begin{tabular}{llcccccccc}
\toprule
Task & Method
& SR$\uparrow$
& MC@25$\uparrow$
& MC@50$\uparrow$
& MC@75$\uparrow$
& MP$\uparrow$
& PPL$\uparrow$
& OPS$\uparrow$
& UPS$\uparrow$ \\
\midrule

\multirow{3}{*}{PackFruit}
& XR-1
& 0.00 & 43.33 & 6.66 & 0.00
& 18.05 & 10.93 & 14.16 & 27.50 \\

& MOST-Raw
& 0.00
& \cellcolor{bestblue!35}76.66
& \cellcolor{bestblue!35}40.00
& 0.00
& \cellcolor{bestblue!35}38.61
& \cellcolor{bestblue!35}23.72
& \cellcolor{bestblue!35}34.16
& \cellcolor{bestblue!35}57.50 \\

& \textbf{MOST (Ours)}
& \cellcolor{bestblue}\textbf{10.00}
& \cellcolor{bestblue}\textbf{93.33}
& \cellcolor{bestblue}\textbf{76.66}
& \cellcolor{bestblue}\textbf{30.00}
& \cellcolor{bestblue}\textbf{61.84}
& \cellcolor{bestblue}\textbf{43.57}
& \cellcolor{bestblue}\textbf{58.84}
& \cellcolor{bestblue}\textbf{82.99} \\

\midrule

\multirow{3}{*}{\shortstack[l]{Prepare\\Breakfast}}
& XR-1
& 0.00
& 66.66
& 0.00
& 0.00
& 26.66
& \cellcolor{bestblue!35}17.93
& \cellcolor{bestblue!35}20.33
& 35.00 \\

& MOST-Raw
& 0.00
& \cellcolor{bestblue!35}76.66
& \cellcolor{bestblue!35}10.00
& 0.00
& \cellcolor{bestblue!35}27.66
& 15.76
& 19.33
& \cellcolor{bestblue!35}42.33 \\

& \textbf{MOST (Ours)}
& \cellcolor{bestblue}\textbf{6.66}
& \cellcolor{bestblue}\textbf{86.66}
& \cellcolor{bestblue}\textbf{16.66}
& \cellcolor{bestblue}\textbf{6.66}
& \cellcolor{bestblue}\textbf{37.14}
& \cellcolor{bestblue}\textbf{36.19}
& \cellcolor{bestblue}\textbf{35.95}
& \cellcolor{bestblue}\textbf{71.66} \\

\midrule

\multirow{3}{*}{\shortstack[l]{CleanAndStore\\AfterMeal}}
& XR-1
& 0.00
& \cellcolor{bestblue}\textbf{33.33}
& 3.33
& 0.00
& 17.77
& 6.87
& 10.27
& 29.16 \\

& MOST-Raw
& 0.00
& \cellcolor{bestblue}\textbf{33.33}
& \cellcolor{bestblue}\textbf{10.00}
& 0.00
& \cellcolor{bestblue!35}20.00
& \cellcolor{bestblue!35}14.00
& \cellcolor{bestblue!35}14.66
& \cellcolor{bestblue!35}38.66 \\

& \textbf{MOST (Ours)}
& \cellcolor{bestblue}\textbf{3.33}
& 20.00
& \cellcolor{bestblue}\textbf{10.00}
& \cellcolor{bestblue}\textbf{6.66}
& \cellcolor{bestblue}\textbf{27.69}
& \cellcolor{bestblue}\textbf{19.32}
& \cellcolor{bestblue}\textbf{21.53}
& \cellcolor{bestblue}\textbf{73.84} \\

\midrule

\multirow{3}{*}{Average}
& XR-1
& 0.00 & 47.77 & 3.33 & 0.00
& 20.83 & 11.91 & 14.92 & 30.55 \\

& MOST-Raw
& 0.00
& \cellcolor{bestblue!35}62.22
& \cellcolor{bestblue!35}20.00
& 0.00
& \cellcolor{bestblue!35}28.75
& \cellcolor{bestblue!35}17.82
& \cellcolor{bestblue!35}22.71
& \cellcolor{bestblue!35}46.16 \\

& \textbf{MOST (Ours)}
& \cellcolor{bestblue}\textbf{6.66}
& \cellcolor{bestblue}\textbf{66.66}
& \cellcolor{bestblue}\textbf{34.44}
& \cellcolor{bestblue}\textbf{14.44}
& \cellcolor{bestblue}\textbf{42.22}
& \cellcolor{bestblue}\textbf{33.03}
& \cellcolor{bestblue}\textbf{38.77}
& \cellcolor{bestblue}\textbf{76.17} \\

\bottomrule
\end{tabular}%
}
\end{table}

We next test horizon effects by sorting the 16 Composite-Seen Target-100\% tasks by demonstration length into four groups. Fig.~\ref{fig:exp_horizon} reports each method's change from XR-1. MOST improves all four metrics in every group. MOST-Raw remains competitive in G2, but the SR, MP, and OPS gaps are largest in G3 and G4. This growing separation suggests that structured task-state memory becomes more useful as the horizon grows.

\subsection{Multi-Minute Evaluation on RoboCasa-Long}
\label{sec:robocasa_long}

\begin{figure*}[t]
    \centering
    \includegraphics[width=\textwidth]{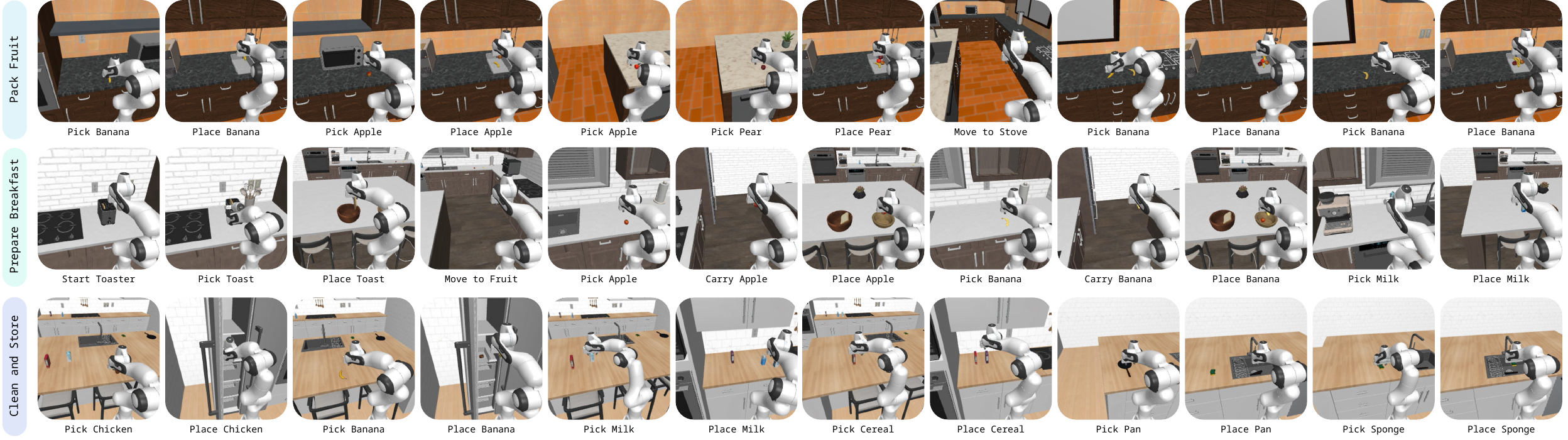}
    \caption{\textbf{RoboCasa-Long task demonstrations.} From top to bottom, each row follows one \emph{PackFruit}, \emph{PrepareBreakfast}, or \emph{CleanAndStoreAfterMeal} demonstration from left to right.}
    \label{fig:robocasa_long_demos}
\end{figure*}

To evaluate policy memory beyond the shorter RoboCasa365 episodes, we construct RoboCasa-Long, a benchmark of three multi-minute, multi-stage mobile manipulation tasks. Each requires the robot to navigate across work areas and execute a sequence of object interactions while retaining earlier progress and persistent changes. We collect 200 demonstrations per task, averaging 3.8 minutes, and evaluate each method over 30 rollouts per task. Fig.~\ref{fig:robocasa_long_demos} visualizes one demonstration from each task and shows the complete navigation and manipulation sequence.

Using matched data and training schedules, we compare memoryless XR-1, visual-history MOST-Raw, and MOST. Tab.~\ref{tab:robocasa_long} reports terminal success and progress-aware performance. MOST-Raw improves average progress over XR-1, while MOST is the only method with nonzero SR on all three tasks and leads every average metric. It leads every metric on \emph{PackFruit} and \emph{PrepareBreakfast}. \emph{CleanAndStoreAfterMeal} remains challenging. MOST-Raw reaches the first milestone more often, but MOST matches MC@50 and leads all remaining completion and progress metrics. These results suggest that task-state memory better sustains progress across long action sequences, although the low absolute SR shows that full-task execution remains challenging.

\begin{table}[t]
\captionsetup{skip=4pt}
\centering
\caption{
Controlled memory-representation study on RoboCasa365 Target-100\% Composite-Seen.
Darker blue denotes higher performance; the lowest result is unshaded and the best is \textbf{bold}.
}
\label{tab:representation_study}

\footnotesize
\setlength{\tabcolsep}{4.0pt}
\renewcommand{\arraystretch}{0.6}

\resizebox{\linewidth}{!}{%
\begin{tabular}{lcccccccc}
\toprule
Memory Representation
& SR$\uparrow$
& MC@25$\uparrow$
& MC@50$\uparrow$
& MC@75$\uparrow$
& MP$\uparrow$
& PPL$\uparrow$
& OPS$\uparrow$
& UPS$\uparrow$ \\
\midrule

No Memory (XR-1)
& \cellcolor{bestblue!17}57.91 & \cellcolor{bestblue!33}90.83 & \cellcolor{bestblue!17}84.37 & \cellcolor{bestblue!20}70.41 & \cellcolor{bestblue!20}79.48 & \cellcolor{bestblue!17}63.55 & \cellcolor{bestblue!17}74.83 & \cellcolor{bestblue!17}84.56 \\

Visual History (MOST-Raw)
& \cellcolor{bestblue!33}63.33 & \cellcolor{bestblue!17}90.62 & \cellcolor{bestblue!50}85.00 & \cellcolor{bestblue!40}73.95 & \cellcolor{bestblue!40}81.68 & \cellcolor{bestblue!33}65.06 & \cellcolor{bestblue!33}76.67 & \cellcolor{bestblue!50}86.82 \\

Current Task State
& 56.45 & 89.37 & 81.66 & 68.54 & 77.50 & 61.35 & 72.24 & 83.00 \\

Task-State History
& \cellcolor{bestblue!50}65.20 & \cellcolor{bestblue!50}91.04 & \cellcolor{bestblue!33}84.79 & \cellcolor{bestblue!60}75.83 & \cellcolor{bestblue!60}82.18 & \cellcolor{bestblue!50}66.23 & \cellcolor{bestblue!50}77.88 & \cellcolor{bestblue!33}86.67 \\

Subtask Trajectory
& \cellcolor{bestblue!83}67.08 & \cellcolor{bestblue!83}92.91 & \cellcolor{bestblue!67}86.45 & \cellcolor{bestblue!80}76.66 & \cellcolor{bestblue!80}83.67 & \cellcolor{bestblue!67}67.88 & \cellcolor{bestblue!67}79.25 & \cellcolor{bestblue!83}88.02 \\

Object-State Trajectory
& \cellcolor{bestblue!67}66.04 & \cellcolor{bestblue!67}92.70 & \cellcolor{bestblue!83}87.08 & \cellcolor{bestblue!80}76.66 & \cellcolor{bestblue!80}83.67 & \cellcolor{bestblue!83}68.09 & \cellcolor{bestblue!83}79.31 & \cellcolor{bestblue!67}86.87 \\

\textbf{Task-State Trajectory (Ours)}
& \cellcolor{bestblue}\textbf{68.75} & \cellcolor{bestblue}\textbf{94.79} & \cellcolor{bestblue}\textbf{89.16} & \cellcolor{bestblue}\textbf{80.20} & \cellcolor{bestblue}\textbf{85.88} & \cellcolor{bestblue}\textbf{68.50} & \cellcolor{bestblue}\textbf{82.09} & \cellcolor{bestblue}\textbf{89.37} \\

\bottomrule
\end{tabular}%
}

\end{table}

\subsection{What Should Task Memory Represent?}
\label{sec:representation_study}

To isolate how memory content and temporal extent affect control, Tab.~\ref{tab:representation_study} evaluates seven variants on Target-100\% Composite-Seen. \emph{No Memory} removes the temporal pathway. \emph{Visual History} retains past observations without task-state supervision or future queries. \emph{Current Task State} predicts the current subtask and relevant object states from only the current observation. \emph{Task-State History} applies the same supervision across observed history without future prediction. The three trajectory variants append future queries and supervise the sequence using subtasks, object states, or both.

Visual History improves SR and most progress measures over No Memory, showing the value of temporal context. Current Task State performs worst across all metrics, indicating that an instantaneous state estimate cannot replace history. Supervising the observed history recovers performance, while future prediction yields further gains. The subtask-only and object-state-only variants show complementary strengths, and their joint Task-State Trajectory is best on every metric. This supports representing memory through both execution progress and persistent changes in the world.

\subsection{Real-World Long-Horizon Manipulation}
\label{sec:real_world}

To test task-state memory under real perception and control, we use a wheeled manipulator whose embodiment closely matches our simulated mobile platform. A single policy is jointly trained on \emph{Pack 1 Banana}, \emph{Pack 2 Bananas}, and \emph{Pack 3 Bananas}. During evaluation, its base remains fixed while the robot rotates its waist between pickup and placement tables, transfers bananas one at a time, and presses the completion button. This setting isolates memory-dependent counting and stopping from navigation failures. We run 10 trials per method and task.

\begin{table}[t]
\captionsetup{skip=4pt}
\centering
\caption{
Real-world long-horizon evaluation on three count-conditioned task variants.
For each instruction, the robot must transfer the requested number of bananas one by one and then press the completion button.
}
\label{tab:real_world}

\footnotesize
\setlength{\tabcolsep}{6.0pt}
\renewcommand{\arraystretch}{0.5}

\resizebox{\linewidth}{!}{%
\begin{tabular}{llcccccc}
\toprule
Task & Method
& SR$\uparrow$
& MC@25$\uparrow$
& MC@50$\uparrow$
& MC@75$\uparrow$
& MP$\uparrow$
& PPL$\uparrow$
\\
\midrule

\multirow{3}{*}{\shortstack[l]{Pack 1 Banana}}
& XR-1
& 0.00
& \cellcolor{bestblue}\textbf{80.00}
& 50.00
& 0.00
& 43.33
& 38.33
\\

& MOST-Raw
& \cellcolor{bestblue}\textbf{40.00}
& 70.00
& 50.00
& \cellcolor{bestblue}\textbf{40.00}
& \cellcolor{bestblue!35}53.33
& \cellcolor{bestblue!35}50.00
\\

& \textbf{MOST (Ours)}
& \cellcolor{bestblue}\textbf{40.00}
& \cellcolor{bestblue}\textbf{80.00}
& \cellcolor{bestblue}\textbf{70.00}
& \cellcolor{bestblue}\textbf{40.00}
& \cellcolor{bestblue}\textbf{63.33}
& \cellcolor{bestblue}\textbf{61.67}
\\

\midrule

\multirow{3}{*}{\shortstack[l]{Pack 2 Bananas}}
& XR-1
& 10.00
& 60.00
& 50.00
& 50.00
& 50.00
& 48.00
\\

& MOST-Raw
& \cellcolor{bestblue!35}40.00
& \cellcolor{bestblue}\textbf{90.00}
& \cellcolor{bestblue!35}70.00
& 50.00
& \cellcolor{bestblue!35}68.00
& \cellcolor{bestblue!35}65.00
\\

& \textbf{MOST (Ours)}
& \cellcolor{bestblue}\textbf{50.00}
& \cellcolor{bestblue}\textbf{90.00}
& \cellcolor{bestblue}\textbf{90.00}
& \cellcolor{bestblue}\textbf{80.00}
& \cellcolor{bestblue}\textbf{82.00}
& \cellcolor{bestblue}\textbf{79.50}
\\

\midrule

\multirow{3}{*}{\shortstack[l]{Pack 3 Bananas}}
& XR-1
& 10.00
& \cellcolor{bestblue!35}80.00
& \cellcolor{bestblue!35}60.00
& 40.00
& \cellcolor{bestblue!35}57.14
& \cellcolor{bestblue!35}54.17
\\

& MOST-Raw
& \cellcolor{bestblue!35}30.00
& 70.00
& 50.00
& \cellcolor{bestblue!35}50.00
& 55.71
& 53.93
\\

& \textbf{MOST (Ours)}
& \cellcolor{bestblue}\textbf{40.00}
& \cellcolor{bestblue}\textbf{100.00}
& \cellcolor{bestblue}\textbf{80.00}
& \cellcolor{bestblue}\textbf{70.00}
& \cellcolor{bestblue}\textbf{81.43}
& \cellcolor{bestblue}\textbf{78.10}
\\

\midrule

\multirow{3}{*}{Average}
& XR-1
& 6.66
& 73.33
& 53.33
& 30.00
& 50.15
& 46.83
\\

& MOST-Raw
& \cellcolor{bestblue!35}36.66
& \cellcolor{bestblue!35}76.66
& \cellcolor{bestblue!35}56.66
& \cellcolor{bestblue!35}46.66
& \cellcolor{bestblue!35}59.01
& \cellcolor{bestblue!35}56.31
\\

& \textbf{MOST (Ours)}
& \cellcolor{bestblue}\textbf{43.33}
& \cellcolor{bestblue}\textbf{90.00}
& \cellcolor{bestblue}\textbf{80.00}
& \cellcolor{bestblue}\textbf{63.33}
& \cellcolor{bestblue}\textbf{75.58}
& \cellcolor{bestblue}\textbf{73.09}
\\

\bottomrule
\end{tabular}%
}

\end{table}

\begin{figure}[t]
    \centering
    \captionsetup{skip=5pt}
    \includegraphics[width=\linewidth]{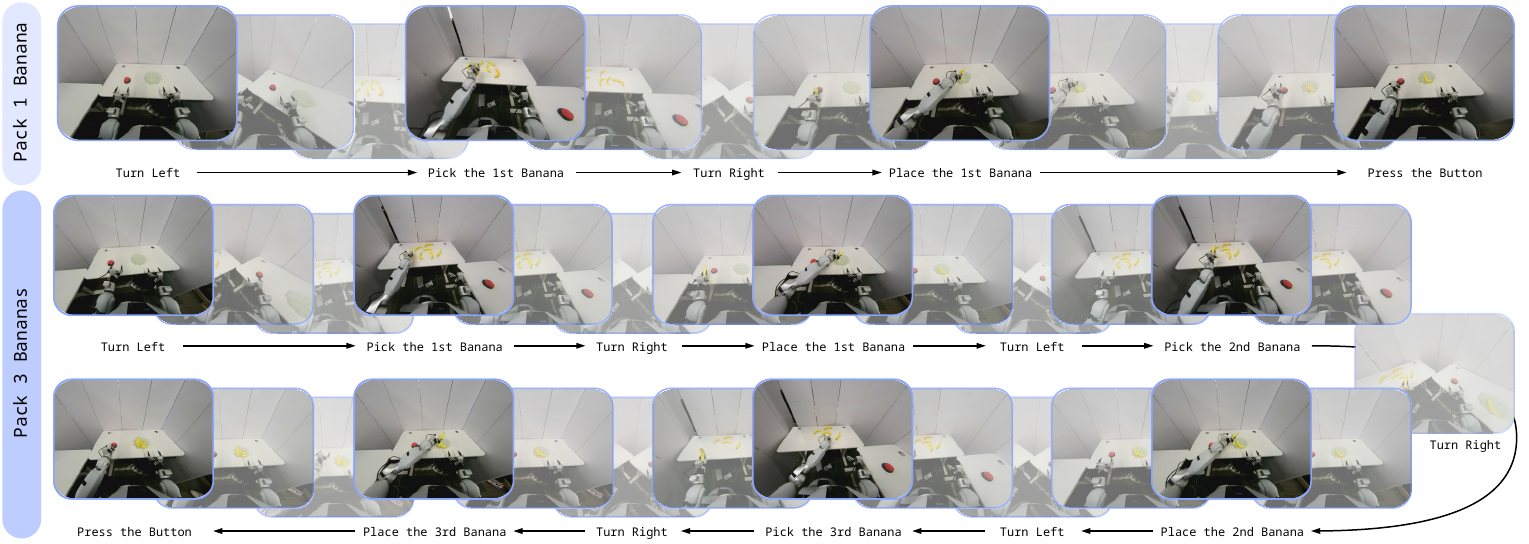}
    \caption{\textbf{Representative successful real-world rollouts on a wheeled manipulator.} MOST transfers one banana and stops in the top row. For \emph{Pack 3 Bananas}, the trajectory continues from the middle row into the bottom row, which is read from right to left, before the robot presses the completion button.}
    \label{fig:real_world_rollouts}
\end{figure}

In Tab.~\ref{tab:real_world}, visual history raises average SR from 6.66\% for XR-1 to 36.66\% for MOST-Raw, while MOST reaches 43.33\%. MOST ties MOST-Raw on \emph{Pack 1 Banana}, improves SR by 10 points on both longer variants, and raises MP and PPL on every task. These results show that task-state memory helps track repetitions and stop correctly. Given only 10 trials per task, we interpret SR with the progress metrics. Fig.~\ref{fig:real_world_rollouts} illustrates this behavior. MOST stops after one placement for the one-banana instruction, but maintains progress through repeated waist rotations and transfers before stopping after the third placement for \emph{Pack 3 Bananas}.

\section{Conclusion}
We introduced MOST, an end-to-end policy that represents internal memory as the evolution of object states and task progress rather than as unstructured visual history. Historical tokens summarize observed task state, while future queries encourage the memory to anticipate its subsequent evolution and provide context for action generation. Results on RoboCasa365, our newly introduced RoboCasa-Long benchmark, and count-conditioned manipulation with a Galbot G1 support this representation for long and repetitive tasks. The current study relies on closed-set state supervision, and full-task success remains limited on the longest tasks. The real-world evaluation also isolates memory under a fixed-base setting. Future work should learn more open-ended task-state abstractions and evaluate MOST on tasks that require navigation, manipulation, and longer-term adaptation.


\subsection*{AI Use Statement}

Generative AI tools were used to assist with drafting and polishing the writing of this manuscript. The authors directed the content, reviewed and revised all AI-assisted text, and take full responsibility for the final paper.

\subsection*{Reproducibility Statement}

Implementation and optimization details, benchmark construction,
task-state annotations, evaluation protocols, and complete metric
definitions are provided in the appendix.


\bibliography{references}
\bibliographystyle{iclr2027_conference}


\appendix
\newpage

\section{Implementation Details}
\label{app:implementation}

\subsection{Model Configuration}
The Memory Expert uses 384-dimensional tokens and 36 layers aligned with the action network. It has six self-attention heads, eight view-fusion heads, and eight heads in each memory-to-action attention layer. We use a block size of $B=8$, $K=16$ future queries, and at most 128 instruction tokens. The memory image encoder is initialized from DINOv2 ViT-S/14~\citep{oquab2023dinov2}. Three $224\!\times\!224$ views are encoded independently, and a learned query fuses their global features into one token at each sampled timestep. During training, gradients update the image encoder through the newest memory observation, while earlier history frames are encoded without image-encoder gradients. At inference, completed eight-token blocks are cached. The current block and future queries are updated at each replanning step, and past RGB frames are discarded after encoding. MOST-Raw retains the historical visual tokens and memory-to-action pathway, but removes the future queries and task-state supervision.

\subsection{Optimization}
All compared variants use the same benchmark-specific initialization, data, and optimization schedule. We train with BF16 mixed precision and DeepSpeed using FusedAdam with $\beta_1=0.9$, $\beta_2=0.95$, weight decay $0.1$, and gradient-norm clipping at $1.0$. The learning rate increases linearly from $5\!\times\!10^{-7}$ to $2\!\times\!10^{-5}$ over 500 warmup steps and then follows a cosine schedule to $5\!\times\!10^{-6}$. We use an effective global batch size of 128 and random seed 42. The subtask and object-state loss weights are both set to $0.25$ for RoboCasa365, $1.0$ for RoboCasa-Long, and $0.75$ for the real-world experiments. RoboCasa365 finetuning runs for 60k steps. RoboCasa-Long uses only the three Long tasks and is trained for 20k steps without RoboCasa365 replay. Real-world training also runs for 20k steps. Training is distributed over eight NVIDIA A800 80\,GB GPUs.

\subsection{Visual and Temporal Inputs}
RoboCasa365 and RoboCasa-Long use the left agent view, right agent view, and wrist view at a control frequency of 20\,Hz. The current-observation pathway receives four frames separated by two source frames and predicts 16-step action chunks. The memory pathway samples one multi-view observation every 16 frames, which gives one memory token every 0.8\,s. The real-world policy uses the left and right head cameras and the left-arm camera at 30\,Hz. Its current images are resized to $448\!\times\!320$, while memory images are resized to $224\!\times\!224$. It predicts 30-step action chunks and samples memory every 30 frames, which gives one memory token per second.

\section{Benchmarks and Datasets}
\label{app:experimental_details}

\subsection{RoboCasa365}

We use the 50-task RoboCasa365 target suite~\citep{nasiriany2026robocasa365}, which contains 18 Atomic, 16 Composite-Seen, and 16 Composite-Unseen tasks. All policies start from the same checkpoint pretrained on Human300, and each target split is finetuned independently. Target-10\% and Target-100\% use fixed episode subsets containing 50 and 500 demonstrations per task, respectively. We train each target setting for 60k steps and evaluate 30 rollouts per task. Evaluation uses the target kitchens and the task-specific horizons registered by RoboCasa365. A rollout terminates when the simulator reports success, termination, truncation, or the maximum horizon. The overall results average the three splits according to their 18, 16, and 16 tasks.

\subsection{RoboCasa-Long}

We introduce RoboCasa-Long to evaluate memory on multi-minute mobile manipulation beyond the shorter RoboCasa365 episodes. It contains \emph{PackFruit}, \emph{PrepareBreakfast}, and \emph{CleanAndStoreAfterMeal}. Their language instructions are:
\begin{itemize}
    \item \textbf{PackFruit:} ``Collect and pack all the fruit from the counter near the microwave, the dining counter, and the counter near the stove into the tupperware.''
    \item \textbf{PrepareBreakfast:} ``Toast the bread and put it in the bowl on the left, put the milk beside that bowl, and put the apple and banana in the bowl on the right.''
    \item \textbf{CleanAndStoreAfterMeal:} ``Put the chicken drumstick and banana in the tupperware on the second shelf of the freezer, put the cereal box and milk carton beside their matching items on the counter by the stove, and put the pan and sponge in the island sink.''
\end{itemize}
Together, these tasks require repeated object transfers, persistent tracking of changes that leave the current view, and navigation among several kitchen work areas. Representative demonstrations are shown in Fig.~\ref{fig:robocasa_long_demos}.

We collect 200 demonstrations for each task. Their average durations are approximately 4.1, 3.6, and 3.8 minutes for \emph{PackFruit}, \emph{PrepareBreakfast}, and \emph{CleanAndStoreAfterMeal}, respectively. The three tasks are trained jointly for 20k steps using only RoboCasa-Long data, without replay from RoboCasa365. We evaluate 30 rollouts per task under the same task environments used for data collection. Success is determined by each environment's complete task success condition rather than by reaching an intermediate stage.

\subsection{Real-World Setup}

Our real-world evaluation uses a Galbot G1 in a fixed-base setup with a pickup table and a placement table. The robot rotates its waist to switch between the two work areas and uses its left arm and gripper to transfer bananas into a bowl before pressing a completion button. We jointly train one instruction-conditioned policy on \emph{Pack 1 Banana}, \emph{Pack 2 Bananas}, and \emph{Pack 3 Bananas}. Their instructions request the corresponding number of transfers followed by a button press.

The training set contains 180 demonstrations in LeRobot format, with 60 episodes for each task. The dataset records four camera streams, while training uses the left and right head cameras and the left-arm camera. We evaluate each method in 10 trials per task. A rollout is successful only when the robot transfers exactly the requested number of bananas into the bowl and then successfully presses the button. Incorrect counts, incomplete transfers, premature termination, and a missing or failed button press are counted as failures. Progress is scored from the recorded trials using the same predefined task stages for all methods.

\subsection{Task-State Annotation Format}

Each episode is paired with a JSON file containing \texttt{action\_config} and \texttt{state\_config}. For RoboCasa365 and RoboCasa-Long, these annotations are generated automatically from privileged simulator information. The real-world data are annotated manually using the same schema. \texttt{action\_config} divides the episode into inclusive frame intervals. Each entry contains \texttt{start\_frame}, \texttt{end\_frame}, an \texttt{action\_text} describing the active subtask, and a coarser \texttt{skill} type such as \texttt{Pick}, \texttt{Move}, or \texttt{Place}. We use \texttt{action\_text}, rather than \texttt{skill}, as the subtask prediction target.

\texttt{state\_config} lists only the task-relevant objects rather than every object in the scene. Each object entry contains \texttt{object\_id}, \texttt{category}, \texttt{instance\_id}, and a list of inclusive intervals paired with \texttt{state\_text}. Distinct instance indices allow multiple objects from the same category to receive separate state targets. Object-state boundaries follow physical state changes and are not forced to coincide with subtask boundaries. For example, a pick subtask may continue while the arm retracts after a successful grasp, even though the object's state has already changed to \texttt{in hand}.

A complete annotation example is shown below.

{\scriptsize
\begin{verbatim}
{
  "episode_id": "000000",
  "label_info": {
    "action_config": [
      {
        "start_frame": 0,
        "end_frame": 78,
        "action_text":
          "Pick up the kettle from the counter",
        "skill": "Pick"
      },
      {
        "start_frame": 79,
        "end_frame": 200,
        "action_text":
          "Place the kettle on the tray",
        "skill": "Place"
      },
      {
        "start_frame": 201,
        "end_frame": 375,
        "action_text":
          "Pick up the mug from the cabinet",
        "skill": "Pick"
      },
      {
        "start_frame": 376,
        "end_frame": 503,
        "action_text":
          "Place the mug on the tray",
        "skill": "Place"
      },
      {
        "start_frame": 504,
        "end_frame": 970,
        "action_text":
          "Close the cabinet doors",
        "skill": "Close"
      }
    ],
    "state_config": [
      {
        "object_id": "mug_0",
        "category": "mug",
        "instance_id": 0,
        "state": [
          {
            "start_frame": 0,
            "end_frame": 374,
            "state_text": "in the cabinet"
          },
          {
            "start_frame": 375,
            "end_frame": 503,
            "state_text": "in hand"
          },
          {
            "start_frame": 504,
            "end_frame": 970,
            "state_text": "on the tray"
          }
        ]
      },
      {
        "object_id": "kettle_0",
        "category": "kettle",
        "instance_id": 0,
        "state": [
          {
            "start_frame": 0,
            "end_frame": 77,
            "state_text": "on the counter"
          },
          {
            "start_frame": 78,
            "end_frame": 200,
            "state_text": "in hand"
          },
          {
            "start_frame": 201,
            "end_frame": 970,
            "state_text": "on the tray"
          }
        ]
      },
      {
        "object_id": "cabinet doors_0",
        "category": "cabinet doors",
        "instance_id": 0,
        "state": [
          {
            "start_frame": 0,
            "end_frame": 954,
            "state_text": "open"
          },
          {
            "start_frame": 955,
            "end_frame": 970,
            "state_text": "closed"
          }
        ]
      }
    ]
  }
}
\end{verbatim}
}

For a sampled frame, the active \texttt{action\_text} supplies its subtask target, and the active \texttt{state\_text} of every listed object supplies the corresponding object-state targets. Historical observations are sampled from the episode prefix at interval $\Delta$, where $\Delta=16$ in simulation and $\Delta=30$ for the real-world data. For a current frame $t$, future query $k$ is supervised using the labels at $t+k\Delta$. Positions outside the episode or without a valid interval are ignored by the auxiliary losses. The subtask, object-state, and object-category vocabularies are constructed from the unique labels in the selected training data. None of these labels or object identifiers is required at inference.

\subsection{Metrics}
\label{app:metrics}

A progress curve records how much of a task has been achieved throughout one rollout. For a task with $N$ predefined milestones, let $m_j(t)\in\{0,1\}$ indicate whether milestone $j$ is satisfied at control step $t$. We define the normalized progress as
\begin{equation}
    p_t=\frac{1}{N}\sum_{j=1}^{N}m_j(t),
    \qquad p_t\in[0,1].
    \label{eq:simulator_progress}
\end{equation}
Completing one new milestone therefore increases progress by $1/N$ when the remaining milestone states do not change. If a reversible milestone becomes unsatisfied, progress decreases by the same amount. In simulation, the milestone indicators are read directly from task-specific predicates using privileged simulator state at every control step. The curve remains constant between milestone transitions and is therefore staircase-shaped. When the simulator reports full task success, its final value is set to $1$. For real-world trials, we obtain the same discrete representation by manually scoring the predefined milestones from the recorded rollout.

This construction differs from PRM-as-a-Judge 1.5~\citep{liu2026prmjudge}, which uses a learned progress reward model to estimate a dense curve from rollout video before post-processing the prediction. We do not use a learned judge. The input to our metric computation is the discrete curve defined above, obtained from simulator predicates or manual annotation, and the reported curve remains staircase-shaped. The resulting metric values are therefore not directly comparable with those computed from PRM-predicted curves.

Across $R$ rollouts, let $p^{(r)}_{0:T_r}$ be the progress curve of rollout $r$, where $T_r$ is its final curve index. Let $y_r\in\{0,1\}$ be its task-success indicator, $L_r$ its executed control steps, and $H_r$ its maximum rollout horizon. For subtask-based metrics, let $M_r$ be the number of required subtasks, $c_r^{\mathrm{ord}}$ the number completed in the required order, and $c_r^{\mathrm{any}}$ the number completed regardless of order. The notation $\mathbb{I}[\cdot]$ denotes an indicator function. We compute the metrics as follows.

\begin{itemize}
    \item \textbf{Success Rate (SR)} is the fraction of rollouts that satisfy the complete task success condition:
    \begin{equation}
        \mathrm{SR}=\frac{1}{R}\sum_{r=1}^{R}y_r.
    \end{equation}

    \item \textbf{Ordered Progress Score (OPS)} is the fraction of required subtasks completed in the prescribed order, $\mathrm{OPS}^{(r)}=c_r^{\mathrm{ord}}/M_r$.

    \item \textbf{Unordered Progress Score (UPS)} is the fraction completed without considering order, $\mathrm{UPS}^{(r)}=c_r^{\mathrm{any}}/M_r$.

    \item \textbf{Normalized Completion Time (NCT)} measures the fraction of the rollout budget used. Successful rollouts use their actual completion time, while failures are assigned the full budget:
    \begin{equation}
        \mathrm{NCT}^{(r)}=
        \begin{cases}
            \min\!\left(1,L_r/H_r\right), & y_r=1,\\
            1, & y_r=0.
        \end{cases}
    \end{equation}
    Lower NCT indicates faster completion.

    \item \textbf{Max Progress (MP)} is the furthest progress reached during a rollout, $\mathrm{MP}^{(r)}=\max_{0\le t\le T_r}p_t^{(r)}$.

    \item \textbf{Milestone Coverage (MC@$q$)} measures the fraction of rollouts that reach progress threshold $q$. For $q\in\{25,50,75\}$,
    \begin{equation}
        \mathrm{MC@}q=\frac{1}{R}\sum_{r=1}^{R}
        \mathbb{I}\!\left[\mathrm{MP}^{(r)}\ge q/100\right].
    \end{equation}

    \item \textbf{Path-weighted Progress Length (PPL)} rewards reaching high progress through a direct path:
    \begin{equation}
        \mathrm{PPL}^{(r)}=
        \frac{\bigl(\mathrm{MP}^{(r)}\bigr)^2}
        {\sum_{t=1}^{T_r}\left|p_t^{(r)}-p_{t-1}^{(r)}\right|}.
        \label{eq:ppl_metric}
    \end{equation}
    PPL is set to zero when the denominator is zero. A monotonic curve has PPL equal to MP, while regression and repeated recovery reduce PPL.

    \item \textbf{Cumulative Regret Area (CRA)} measures the severity and duration of regression below the best progress achieved so far. Let $b_t^{(r)}=\max_{0\le i\le t}p_i^{(r)}$. Then
    \begin{equation}
        \mathrm{CRA}^{(r)}=
        \frac{1}{(T_r+1)\mathrm{MP}^{(r)}}
        \sum_{t=0}^{T_r}\left(b_t^{(r)}-p_t^{(r)}\right).
    \end{equation}
    We define CRA $=0$ for a zero-progress rollout, for which both the numerator and MP are zero. Lower CRA indicates less severe or persistent regression.

    \item \textbf{Drawdown Recovery Ratio (DRR)} measures recovery after the largest loss of progress. Define the drawdown $d_t^{(r)}=b_t^{(r)}-p_t^{(r)}$ and let $t_r^\star\in\arg\max_t d_t^{(r)}$ be the step of maximum drawdown. For rollouts with $d_{t_r^\star}^{(r)}>0$,
    \begin{equation}
        \mathrm{DRR}^{(r)}=
        \min\!\left(
            1,
            \frac{\max_{t_r^\star\le t\le T_r}p_t^{(r)}-p_{t_r^\star}^{(r)}}
            {d_{t_r^\star}^{(r)}}
        \right).
    \end{equation}
    DRR is averaged only over rollouts that experience a drawdown. A value of $1$ indicates full recovery. If no rollout for a task experiences a drawdown, DRR is undefined and reported as ``--''. This means that no progress regression was observed and should not be interpreted as poor recovery.

    \item \textbf{Failure Near-Success (FNS)} measures how close a failed rollout came to completion:
    \begin{equation}
        \mathrm{FNS}^{(r)}=
        0.5\,\mathrm{MP}^{(r)}
        +0.3\,\mathbb{I}\!\left[\mathrm{MP}^{(r)}\ge0.75\right]
        +0.2\,\mathbb{I}\!\left[\mathrm{MP}^{(r)}\ge0.50\right].
    \end{equation}
    FNS is averaged only over failed rollouts. Higher FNS indicates failure at a later task stage.
\end{itemize}

The main paper reports SR, OPS, UPS, MP, MC@25/50/75, and PPL. The appendix additionally reports NCT, CRA, DRR, and FNS. Unless a table explicitly averages task-level rows, we use the arithmetic mean over all eligible rollouts in the reported set, rather than the median aggregation used by PRM-as-a-Judge 1.5 for continuous metrics. DRR includes only rollouts with a drawdown, and FNS includes only failed rollouts. Higher defined values are better for all reported metrics except NCT and CRA. Scores are expressed as percentages in the tables and truncated to the displayed precision rather than rounded.

\section{More Results}

Tab.~\ref{tab:robocasa365_cs_t10_taskwise} and Tab.~\ref{tab:robocasa365_cu_t10_taskwise} report per-task results for the Target-10\% Composite-Seen and Composite-Unseen settings. Tab.~\ref{tab:robocasa365_cs_t100_taskwise} and Tab.~\ref{tab:robocasa365_cu_t100_taskwise} report the corresponding Target-100\% results.

\begin{table*}[t]
\captionsetup{skip=4pt}
\centering
\caption{Per-task performance (\%) of MOST on RoboCasa365 Target-10\% Composite-Seen.}
\label{tab:robocasa365_cs_t10_taskwise}

\scriptsize
\setlength{\tabcolsep}{2.6pt}
\renewcommand{\arraystretch}{0.95}

\resizebox{\textwidth}{!}{%
\begin{tabular}{lcccccccccccc}
\toprule
Task
& SR$\uparrow$
& OPS$\uparrow$
& UPS$\uparrow$
& NCT$\downarrow$
& MP$\uparrow$
& MC@25$\uparrow$
& MC@50$\uparrow$
& MC@75$\uparrow$
& PPL$\uparrow$
& CRA$\downarrow$
& DRR$\uparrow$
& FNS$\uparrow$ \\
\midrule

DeliverStraw & 33.33 & 50.00 & 75.55 & 86.80 & 68.88 & 83.33 & 73.33 & 50.00 & 50.81 & 11.05 & 25.00 & 46.16 \\
GetToastedBread & 83.33 & 90.83 & 95.83 & 68.03 & 90.83 & 93.33 & 93.33 & 90.00 & 89.72 & 0.00 & 100.00 & 46.50 \\
KettleBoiling & 70.00 & 81.11 & 86.66 & 67.96 & 83.33 & 90.00 & 90.00 & 70.00 & 83.33 & 0.00 & -- & 35.55 \\
LoadDishwasher & 66.66 & 78.66 & 90.00 & 73.75 & 84.66 & 93.33 & 90.00 & 76.66 & 62.40 & 6.94 & 77.27 & 50.00 \\
PackIdenticalLunches & 20.00 & 47.50 & 69.16 & 92.56 & 67.50 & 100.00 & 76.66 & 43.33 & 41.96 & 18.87 & 62.20 & 52.60 \\
PreSoakPan & 53.33 & 86.66 & 89.33 & 76.38 & 87.33 & 96.66 & 93.33 & 93.33 & 81.50 & 1.12 & 100.00 & 79.28 \\
PrepareCoffee & 13.33 & 33.33 & 60.00 & 93.92 & 50.00 & 96.66 & 40.00 & 13.33 & 40.55 & 34.21 & 6.25 & 27.30 \\
RinseSinkBasin & 56.66 & 70.83 & 70.83 & 74.22 & 70.83 & 76.66 & 76.66 & 73.33 & 70.83 & 0.00 & -- & 37.11 \\
ScrubCuttingBoard & 40.00 & 78.88 & 78.88 & 88.43 & 78.88 & 100.00 & 96.66 & 40.00 & 77.55 & 0.21 & 100.00 & 51.29 \\
SearingMeat & 20.00 & 40.66 & 73.33 & 87.01 & 52.00 & 50.00 & 50.00 & 40.00 & 39.06 & 42.77 & 30.43 & 35.00 \\
SetUpCuttingStation & 63.33 & 82.50 & 88.33 & 68.17 & 85.83 & 96.66 & 96.66 & 86.66 & 63.40 & 3.24 & 90.00 & 67.95 \\
StackBowlsCabinet & 86.66 & 93.33 & 96.66 & 40.83 & 96.66 & 100.00 & 100.00 & 100.00 & 67.05 & 5.37 & 94.73 & 87.50 \\
SteamInMicrowave & 53.33 & 80.55 & 88.33 & 85.21 & 81.66 & 86.66 & 86.66 & 80.00 & 62.87 & 5.84 & 91.66 & 61.78 \\
StirVegetables & 16.66 & 81.66 & 83.33 & 92.71 & 82.22 & 96.66 & 93.33 & 93.33 & 48.09 & 3.12 & 96.66 & 85.33 \\
StoreLeftoversInBowl & 90.00 & 93.33 & 96.11 & 55.53 & 96.11 & 96.66 & 96.66 & 96.66 & 52.96 & 3.23 & 96.66 & 63.88 \\
WashLettuce & 50.00 & 60.00 & 65.83 & 72.87 & 60.00 & 70.00 & 66.66 & 53.33 & 60.00 & 0.00 & -- & 18.66 \\

\midrule
Average & 51.04 & 71.86 & 81.76 & 76.52 & 77.29 & 89.16 & 82.50 & 68.75 & 62.00 & 8.50 & 74.68 & 52.87 \\

\bottomrule
\end{tabular}%
}

\end{table*}

\begin{table*}[t]
\captionsetup{skip=4pt}
\centering
\caption{Per-task performance (\%) of MOST on RoboCasa365 Target-10\% Composite-Unseen.}
\label{tab:robocasa365_cu_t10_taskwise}

\scriptsize
\setlength{\tabcolsep}{2.6pt}
\renewcommand{\arraystretch}{0.95}

\resizebox{\textwidth}{!}{%
\begin{tabular}{lcccccccccccc}
\toprule
Task
& SR$\uparrow$
& OPS$\uparrow$
& UPS$\uparrow$
& NCT$\downarrow$
& MP$\uparrow$
& MC@25$\uparrow$
& MC@50$\uparrow$
& MC@75$\uparrow$
& PPL$\uparrow$
& CRA$\downarrow$
& DRR$\uparrow$
& FNS$\uparrow$ \\
\midrule

ArrangeBreadBasket & 86.66 & 88.00 & 94.83 & 40.12 & 89.50 & 93.33 & 86.66 & 86.66 & 51.16 & 2.57 & 96.42 & 10.62 \\
ArrangeTea & 60.00 & 80.66 & 89.33 & 85.53 & 85.33 & 100.00 & 93.33 & 73.33 & 72.97 & 3.45 & 57.14 & 58.33 \\
BreadSelection & 50.00 & 72.50 & 79.16 & 76.79 & 75.83 & 100.00 & 93.33 & 60.00 & 47.06 & 5.90 & 87.50 & 49.16 \\
CategorizeCondiments & 13.33 & 40.83 & 73.33 & 96.22 & 55.83 & 100.00 & 76.66 & 33.33 & 37.69 & 21.05 & 69.56 & 46.05 \\
CuttingToolSelection & 63.33 & 68.33 & 80.00 & 64.37 & 73.33 & 83.33 & 83.33 & 63.33 & 42.50 & 3.91 & 85.00 & 24.54 \\
GarnishPancake & 80.00 & 87.77 & 93.33 & 64.05 & 92.22 & 100.00 & 96.66 & 80.00 & 82.59 & 3.27 & 55.55 & 47.22 \\
GatherTableware & 40.00 & 60.83 & 60.83 & 86.53 & 64.16 & 76.66 & 76.66 & 53.33 & 64.16 & 0.00 & -- & 39.02 \\
HeatKebabSandwich & 26.66 & 54.44 & 77.22 & 93.83 & 63.88 & 90.00 & 73.33 & 43.33 & 39.78 & 13.06 & 80.95 & 44.92 \\
MakeIceLemonade & 56.66 & 80.00 & 84.66 & 84.28 & 82.00 & 96.66 & 86.66 & 66.66 & 78.72 & 2.36 & 40.00 & 50.00 \\
PanTransfer & 30.00 & 61.11 & 83.33 & 83.50 & 65.55 & 100.00 & 63.33 & 33.33 & 54.48 & 6.33 & 86.36 & 36.34 \\
PortionHotDogs & 16.66 & 52.08 & 79.58 & 95.13 & 65.41 & 100.00 & 73.33 & 43.33 & 27.13 & 17.72 & 91.55 & 52.45 \\
RecycleBottlesByType & 60.00 & 82.77 & 93.88 & 69.11 & 87.77 & 100.00 & 100.00 & 70.00 & 68.61 & 4.25 & 84.37 & 62.22 \\
SeparateFreezerRack & 56.66 & 70.00 & 82.50 & 79.95 & 75.83 & 100.00 & 76.66 & 66.66 & 68.26 & 7.79 & 72.72 & 38.26 \\
WaffleReheat & 66.66 & 89.33 & 93.33 & 61.59 & 90.66 & 100.00 & 100.00 & 86.66 & 90.66 & 0.00 & -- & 74.00 \\
WashFruitColander & 43.33 & 80.47 & 86.19 & 81.44 & 84.28 & 93.33 & 93.33 & 83.33 & 47.22 & 5.94 & 92.85 & 74.95 \\
WeighIngredients & 30.00 & 36.66 & 84.44 & 81.36 & 61.11 & 93.33 & 60.00 & 30.00 & 47.77 & 41.46 & 12.50 & 30.79 \\

\midrule
Average & 48.75 & 69.11 & 83.49 & 77.74 & 75.79 & 95.41 & 83.33 & 60.83 & 57.55 & 8.69 & 72.32 & 46.18 \\

\bottomrule
\end{tabular}%
}

\end{table*}

\begin{table*}[t]
\captionsetup{skip=4pt}
\centering
\caption{Per-task performance (\%) of MOST on RoboCasa365 Target-100\% Composite-Seen.}
\label{tab:robocasa365_cs_t100_taskwise}

\scriptsize
\setlength{\tabcolsep}{2.6pt}
\renewcommand{\arraystretch}{0.95}

\resizebox{\textwidth}{!}{%
\begin{tabular}{lcccccccccccc}
\toprule
Task
& SR$\uparrow$
& OPS$\uparrow$
& UPS$\uparrow$
& NCT$\downarrow$
& MP$\uparrow$
& MC@25$\uparrow$
& MC@50$\uparrow$
& MC@75$\uparrow$
& PPL$\uparrow$
& CRA$\downarrow$
& DRR$\uparrow$
& FNS$\uparrow$ \\
\midrule

DeliverStraw & 73.33 & 81.11 & 94.44 & 67.17 & 92.22 & 96.66 & 96.66 & 83.33 & 66.78 & 7.26 & 65.00 & 64.16 \\
GetToastedBread & 86.66 & 88.33 & 94.16 & 67.99 & 89.16 & 90.00 & 90.00 & 90.00 & 83.20 & 0.72 & 80.00 & 21.87 \\
KettleBoiling & 90.00 & 94.44 & 96.66 & 57.19 & 96.66 & 100.00 & 100.00 & 90.00 & 89.11 & 0.08 & 100.00 & 53.33 \\
LoadDishwasher & 96.66 & 99.33 & 99.33 & 58.80 & 99.33 & 100.00 & 100.00 & 100.00 & 73.32 & 0.23 & 100.00 & 90.00 \\
PackIdenticalLunches & 40.00 & 66.25 & 78.75 & 82.80 & 78.33 & 100.00 & 93.33 & 60.00 & 62.69 & 13.69 & 66.66 & 59.72 \\
PreSoakPan & 73.33 & 90.66 & 93.33 & 75.65 & 92.66 & 96.66 & 96.66 & 96.66 & 86.54 & 1.62 & 66.66 & 80.00 \\
PrepareCoffee & 20.00 & 51.11 & 68.88 & 92.27 & 63.33 & 100.00 & 66.66 & 23.33 & 54.77 & 16.23 & 16.66 & 40.00 \\
RinseSinkBasin & 63.33 & 83.33 & 83.33 & 68.97 & 83.33 & 93.33 & 93.33 & 83.33 & 83.33 & 0.00 & -- & 60.00 \\
ScrubCuttingBoard & 60.00 & 64.44 & 87.77 & 84.59 & 72.22 & 93.33 & 63.33 & 60.00 & 65.66 & 13.66 & 22.22 & 16.94 \\
SearingMeat & 33.33 & 52.00 & 78.00 & 81.78 & 60.00 & 60.00 & 60.00 & 53.33 & 41.59 & 29.07 & 53.84 & 37.00 \\
SetUpCuttingStation & 90.00 & 95.00 & 97.50 & 55.88 & 95.00 & 100.00 & 100.00 & 90.00 & 61.22 & 1.77 & 100.00 & 45.00 \\
StackBowlsCabinet & 93.33 & 96.66 & 98.33 & 31.28 & 98.33 & 100.00 & 100.00 & 100.00 & 68.89 & 2.57 & 100.00 & 87.50 \\
SteamInMicrowave & 70.00 & 91.11 & 93.88 & 86.52 & 91.66 & 96.66 & 96.66 & 93.33 & 75.38 & 2.99 & 95.00 & 77.22 \\
StirVegetables & 46.66 & 85.55 & 88.33 & 82.55 & 85.55 & 96.66 & 93.33 & 90.00 & 50.27 & 1.09 & 100.00 & 78.33 \\
StoreLeftoversInBowl & 93.33 & 96.66 & 98.88 & 49.77 & 98.88 & 100.00 & 100.00 & 100.00 & 55.83 & 2.39 & 96.66 & 91.66 \\
WashLettuce & 70.00 & 77.50 & 78.33 & 62.91 & 77.50 & 93.33 & 76.66 & 70.00 & 77.50 & 0.00 & -- & 16.94 \\

\midrule
Average & 68.75 & 82.09 & 89.37 & 69.13 & 85.88 & 94.79 & 89.16 & 80.20 & 68.50 & 5.83 & 75.90 & 57.48 \\

\bottomrule
\end{tabular}%
}

\end{table*}

\begin{table*}[t]
\captionsetup{skip=4pt}
\centering
\caption{Per-task performance (\%) of MOST on RoboCasa365 Target-100\% Composite-Unseen.}
\label{tab:robocasa365_cu_t100_taskwise}

\scriptsize
\setlength{\tabcolsep}{2.6pt}
\renewcommand{\arraystretch}{0.95}

\resizebox{\textwidth}{!}{%
\begin{tabular}{lcccccccccccc}
\toprule
Task
& SR$\uparrow$
& OPS$\uparrow$
& UPS$\uparrow$
& NCT$\downarrow$
& MP$\uparrow$
& MC@25$\uparrow$
& MC@50$\uparrow$
& MC@75$\uparrow$
& PPL$\uparrow$
& CRA$\downarrow$
& DRR$\uparrow$
& FNS$\uparrow$ \\
\midrule

ArrangeBreadBasket & 86.66 & 89.33 & 91.33 & 40.80 & 90.00 & 90.00 & 86.66 & 86.66 & 53.16 & 1.38 & 96.29 & 12.50 \\
ArrangeTea & 76.66 & 90.66 & 93.33 & 79.05 & 92.00 & 100.00 & 96.66 & 86.66 & 83.86 & 1.32 & 77.77 & 62.85 \\
BreadSelection & 83.33 & 91.66 & 95.00 & 66.05 & 92.50 & 100.00 & 100.00 & 86.66 & 64.37 & 0.68 & 95.65 & 53.50 \\
CategorizeCondiments & 46.66 & 70.00 & 85.83 & 88.36 & 76.66 & 100.00 & 90.00 & 66.66 & 67.36 & 6.93 & 54.16 & 55.62 \\
CuttingToolSelection & 46.66 & 50.00 & 71.66 & 74.73 & 65.00 & 83.33 & 83.33 & 46.66 & 45.83 & 8.01 & 43.75 & 30.93 \\
GarnishPancake & 70.00 & 81.11 & 91.11 & 72.22 & 90.00 & 100.00 & 100.00 & 70.00 & 78.74 & 6.92 & 33.33 & 53.33 \\
GatherTableware & 56.66 & 73.33 & 73.33 & 79.00 & 73.33 & 80.00 & 80.00 & 66.66 & 73.33 & 0.00 & -- & 36.92 \\
HeatKebabSandwich & 23.33 & 67.77 & 82.77 & 96.66 & 74.44 & 93.33 & 86.66 & 63.33 & 48.44 & 9.18 & 86.00 & 65.50 \\
MakeIceLemonade & 56.66 & 84.66 & 91.33 & 89.93 & 86.66 & 96.66 & 90.00 & 80.00 & 82.62 & 2.20 & 60.00 & 66.15 \\
PanTransfer & 63.33 & 80.00 & 93.33 & 58.66 & 83.33 & 100.00 & 86.66 & 63.33 & 77.96 & 6.95 & 75.00 & 40.00 \\
PortionHotDogs & 63.33 & 76.66 & 93.33 & 75.17 & 85.83 & 100.00 & 86.66 & 80.00 & 43.08 & 8.86 & 90.22 & 57.04 \\
RecycleBottlesByType & 93.33 & 96.66 & 98.88 & 50.69 & 96.66 & 100.00 & 96.66 & 93.33 & 84.92 & 0.14 & 100.00 & 35.00 \\
SeparateFreezerRack & 56.66 & 73.33 & 86.66 & 81.40 & 81.66 & 100.00 & 90.00 & 80.00 & 71.80 & 10.38 & 50.00 & 60.38 \\
WaffleReheat & 96.66 & 99.33 & 99.33 & 38.83 & 99.33 & 100.00 & 100.00 & 100.00 & 99.33 & 0.00 & -- & 90.00 \\
WashFruitColander & 50.00 & 84.28 & 86.66 & 84.38 & 87.14 & 100.00 & 96.66 & 83.33 & 51.93 & 0.44 & 100.00 & 75.80 \\
WeighIngredients & 66.66 & 66.66 & 92.22 & 54.59 & 80.00 & 100.00 & 73.33 & 66.66 & 67.42 & 28.11 & 28.57 & 24.00 \\

\midrule
Average & 64.79 & 79.71 & 89.13 & 70.66 & 84.66 & 96.45 & 90.20 & 76.25 & 68.38 & 5.72 & 70.76 & 51.22 \\

\bottomrule
\end{tabular}%
}

\end{table*}

\clearpage
\section{Visualizations}
\label{app:visualizations}

\begin{figure}[H]
    \centering
    \includegraphics[width=\linewidth,height=0.78\textheight,keepaspectratio]{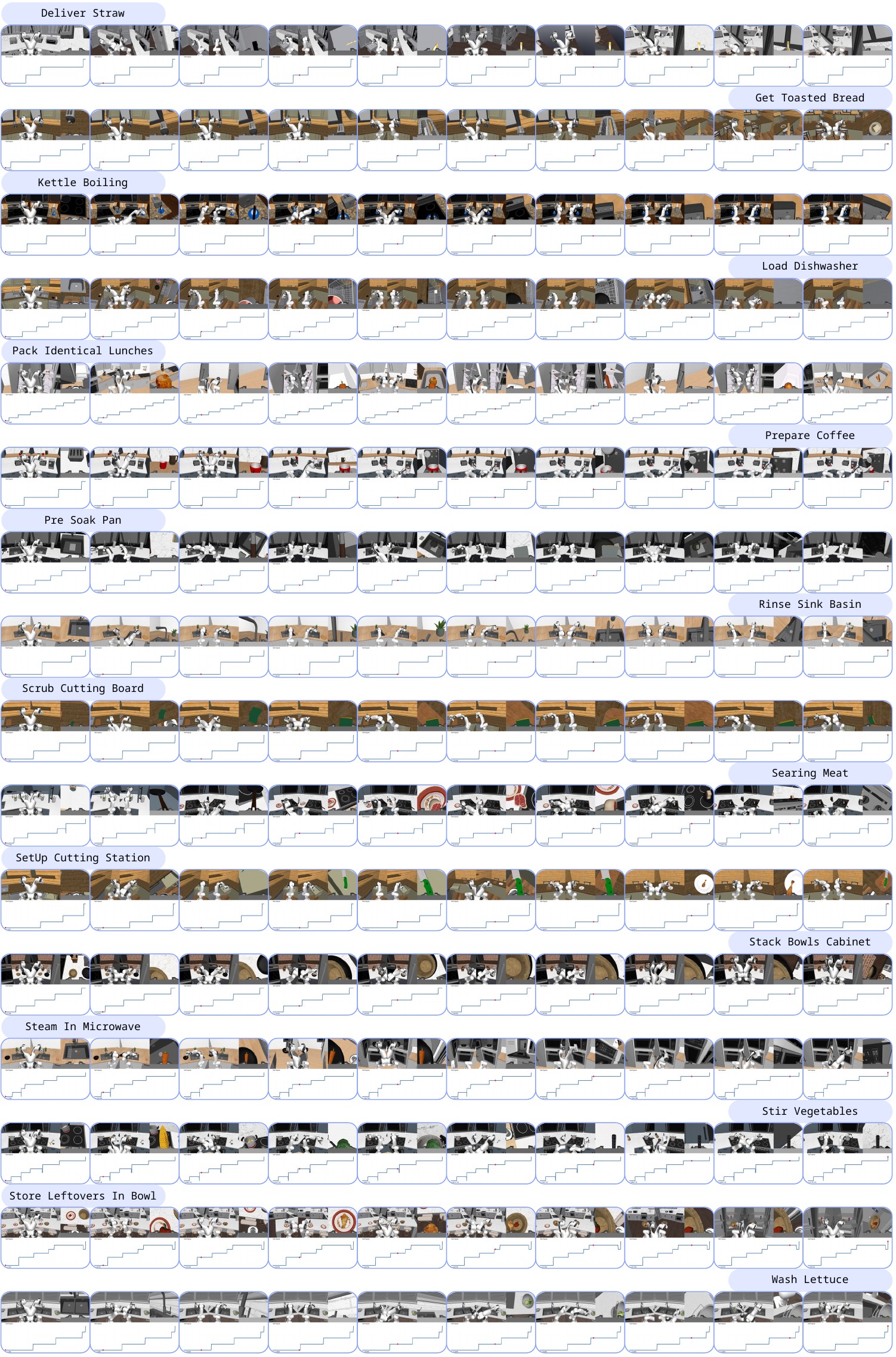}
    \caption{\textbf{RoboCasa365 Composite-Seen visualizations under Target-100\%.} Each row shows ten sampled frames from one rollout of a different task. The staircase beneath each frame traces exact simulator-derived task progress over the rollout, and the red marker indicates the displayed timestep.}
    \label{fig:robocasa365_cs_t100_visualizations}
\end{figure}

\begin{figure}[H]
    \centering
    \includegraphics[width=\linewidth,height=0.86\textheight,keepaspectratio]{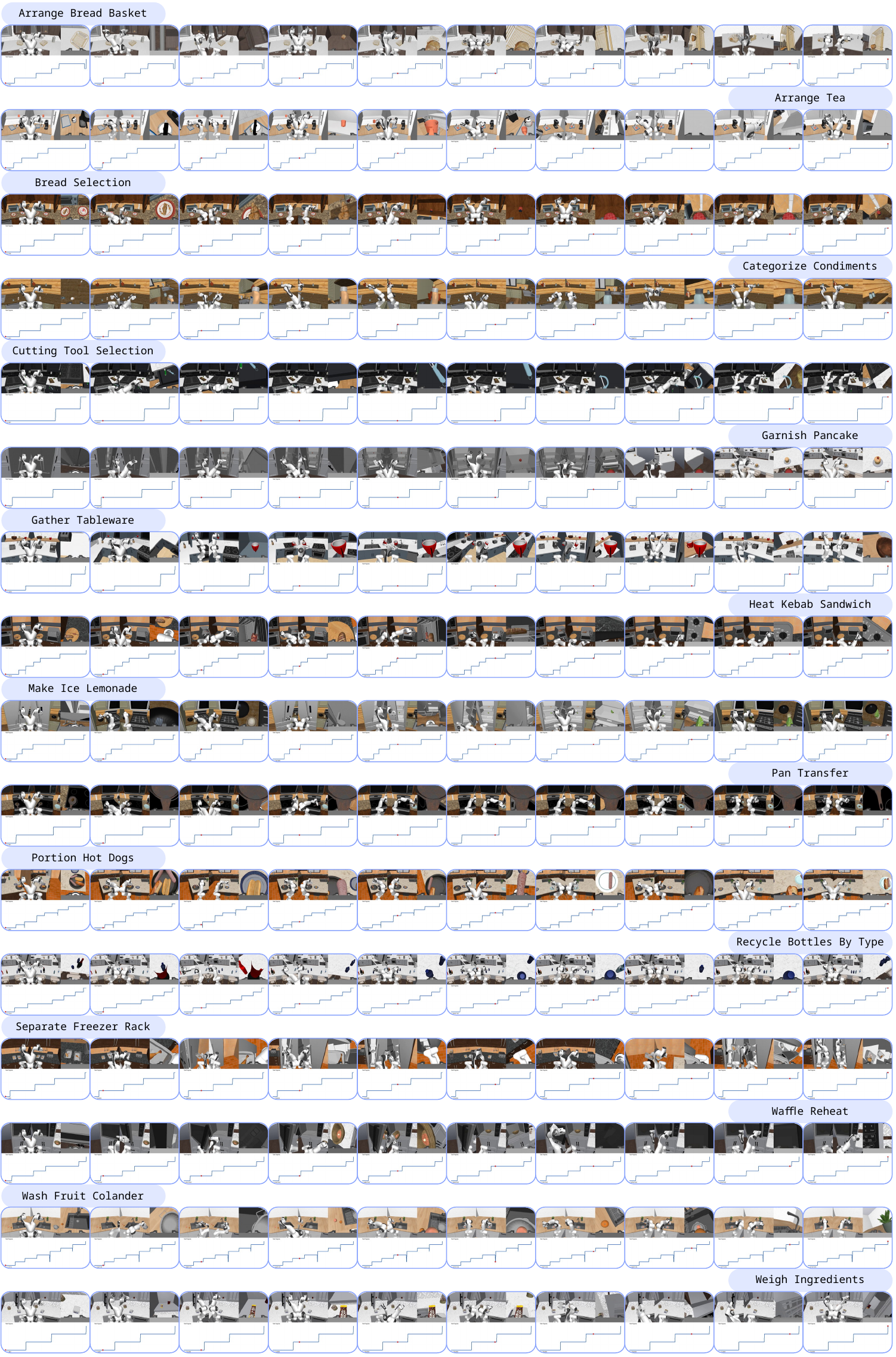}
    \caption{\textbf{RoboCasa365 Composite-Unseen visualizations under Target-100\%.} Each row shows ten sampled frames from one rollout of a different task. The staircase beneath each frame traces exact simulator-derived task progress over the rollout, and the red marker indicates the displayed timestep.}
    \label{fig:robocasa365_cu_t100_visualizations}
\end{figure}

Fig.~\ref{fig:robocasa365_cs_t100_visualizations} and
Fig.~\ref{fig:robocasa365_cu_t100_visualizations}
visualize evaluation rollouts produced by MOST for all 16 Composite-Seen
and 16 Composite-Unseen tasks under Target-100\%.
Each row follows one rollout across ten sampled timesteps together with
its exact simulator-derived progress trajectory.

\clearpage

\end{document}